\pdfoutput=1
\documentclass[10pt,twocolumn,letterpaper]{article}

\usepackage[pagenumbers]{cvpr}   

\usepackage{cuted} 
\usepackage{currfile} 
\usepackage{caption} 

\usepackage{amsmath,amssymb,amsfonts,bm,dsfont}
\usepackage{booktabs}
\usepackage{multirow}
\usepackage{makecell}
\usepackage{colortbl}
\usepackage{pifont}
\usepackage{enumitem}
\usepackage{algorithm}
\usepackage{algpseudocode}
\usepackage{listings}
\usepackage{microtype}
\usepackage{xspace}
\usepackage{placeins}

\newif\ifdraftmarks
\draftmarksfalse

\definecolor{phred}{rgb}{0.80,0.05,0.05}
\definecolor{pendingcol}{rgb}{0.62,0.33,0.02}
\definecolor{figphgray}{gray}{0.45}
\definecolor{rowgray}{gray}{0.93}
\definecolor{agentA}{HTML}{5F8F52}
\definecolor{agentB}{HTML}{8467B0}
\definecolor{bothAB}{HTML}{C9772C}

\newcommand{\method}{MAWAM\xspace}
\newcommand{\bench}{MA-Bench\xspace}
\newcommand{\harness}{MA-Task Harness\xspace}
\newcommand{\pizf}{$\pi_{0.5}$\xspace}
\newcommand{\lingbot}{LingBot-VA\xspace}

\newcommand{\cmark}{\ding{51}}
\newcommand{\xmark}{\ding{55}}

\newcommand{\Ecal}{\mathcal{E}}
\newcommand{\Ocal}{\mathcal{O}}
\newcommand{\Wcal}{\mathcal{W}}
\newcommand{\Ucal}{\mathcal{U}}
\newcommand{\Rcal}{\mathcal{R}}
\newcommand{\Lcal}{\mathcal{L}}
\newcommand{\Hcal}{\mathcal{H}}
\newcommand{\Feas}{\mathrm{Feas}}
\newcommand{\Mcollab}{M_{\mathrm{collab}}}
\newcommand{\Scollab}{S_{\mathrm{collab}}}
\newcommand{\SR}{\mathrm{SR}}
\newcommand{\indic}{\mathds{1}}
\newcommand{\CA}{\mathrm{CA}}
\newcommand{\SA}{\mathrm{SA}}
\newcommand{\LN}{\mathrm{LN}}
\newcommand{\FFN}{\mathrm{FFN}}

\newcommand{\para}[1]{\vspace{1pt}\noindent\textbf{#1}\hspace{0.4em}}

\AtBeginDocument{%
  \setlength{\abovedisplayskip}{5pt plus 1pt minus 1pt}%
  \setlength{\belowdisplayskip}{5pt plus 1pt minus 1pt}%
  \setlength{\abovedisplayshortskip}{3pt plus 1pt}%
  \setlength{\belowdisplayshortskip}{4pt plus 1pt}}

\makeatletter
\renewcommand{\theHALG@line}{\thealgorithm.\arabic{ALG@line}}
\makeatother

\makeatletter
\AtEndPreamble{%
    \crefname{appendix}{Appendix}{Appendices}%
    \Crefname{appendix}{Appendix}{Appendices}%
    \crefname{subappendix}{Appendix}{Appendices}%
    \Crefname{subappendix}{Appendix}{Appendices}%
    \g@addto@macro\appendix{%
        \crefalias{section}{appendix}%
        \crefalias{subsection}{subappendix}%
    }%
}
\newcommand{\@supplfallback}[3]{%
    \global\@namedef{r@#1}{{#2}{1}{}{}{}}%
    \global\@namedef{r@#1@cref}{{#3}{[1][1][]1}{}{}{}}%
}
\AtBeginDocument{%
    \@ifundefined{r@sec:supp:def}{%
        \@supplfallback{sec:supp:def}{A}{[appendix][1][]A}%
        \@supplfallback{sec:supp:harness}{B}{[appendix][2][]B}%
        \@supplfallback{tab:bench_compare}{4}{[table][4][]4}%
        \@supplfallback{sec:supp:bench}{C}{[appendix][3][]C}%
        \@supplfallback{sec:supp:model}{D}{[appendix][4][]D}%
        \@supplfallback{sec:supp:baselines}{E}{[appendix][5][]E}%
        \@supplfallback{sec:supp:results}{F}{[appendix][6][]F}%
        \@supplfallback{sec:supp:consistency}{F.3}{[subappendix][3][6]F.3}%
        \@supplfallback{tab:axes}{9}{[table][9][]9}%
        \@supplfallback{sec:supp:real}{G}{[appendix][7][]G}%
        \@supplfallback{sec:supp:more_cases}{H}{[appendix][8][]H}%
    }{}%
}
\makeatother

\definecolor{cvprblue}{rgb}{0.21,0.49,0.74}
\usepackage[pagebackref,breaklinks,colorlinks,allcolors=cvprblue]{hyperref}

\def\confName{CVPR}
\def\confYear{2027}

\makeatletter
\def\@maketitle{
   \newpage
   \null
   \vskip .06in
   \begin{center}
      {\Large \bf \@title \par}
      \vspace*{9pt}
      {
        \large
        \lineskip .42em
        \begin{tabular}[t]{c}
          \@author
        \end{tabular}
        \par
      }
      \vspace*{6pt}
   \end{center}
}
\makeatother

\title{MAWAM: Joint Imagination for Collaborative Robot Manipulation \\ with a Multi-Agent World Action Model}

\author{%
Rongze Tang\textsuperscript{1,3},\hspace{0.45em}
Chenyu Zhou\textsuperscript{2,3},\hspace{0.45em}
Jianjie Fang\textsuperscript{4},\hspace{0.45em}
Ziming Liu\textsuperscript{4},\hspace{0.45em}
Zhaolu Wang\textsuperscript{3},\hspace{0.45em}
Ziyou Wang\textsuperscript{4},\\[1.5pt]
Xvyuan Liu\textsuperscript{4},\hspace{0.45em}
Wei Wu\textsuperscript{5},\hspace{0.45em}
Chen Gao\textsuperscript{3,4,$\dagger$},\hspace{0.45em}
Yong Li\textsuperscript{4},\hspace{0.45em}
Zhibo Chen\textsuperscript{1,3,$\dagger$}\\[3.5pt]
{\normalsize \textsuperscript{1}University of Science and Technology of China \quad \textsuperscript{2}Nanjing University \quad \textsuperscript{3}Zhongguancun Academy}\\[1pt]
{\normalsize \textsuperscript{4}Tsinghua University \quad \textsuperscript{5}Manifold AI}\\[1.5pt]
{\tt\small chgao96@gmail.com, chenzhibo@ustc.edu.cn}\hspace{0.9em}{\small $^{\dagger}$Corresponding authors}\\[1.5pt]
{\normalsize \textbf{Project Page:}~\href{https://embodiedcity.github.io/MAWAM/}{\textcolor{magenta}{\texttt{https://embodiedcity.github.io/MAWAM/}}}}%
}

\begin{document}
\twocolumn[{%
\renewcommand\twocolumn[1][]{#1}%
\maketitle
\begin{center}
\vspace{-0.8em}
\includegraphics[width=0.985\textwidth]{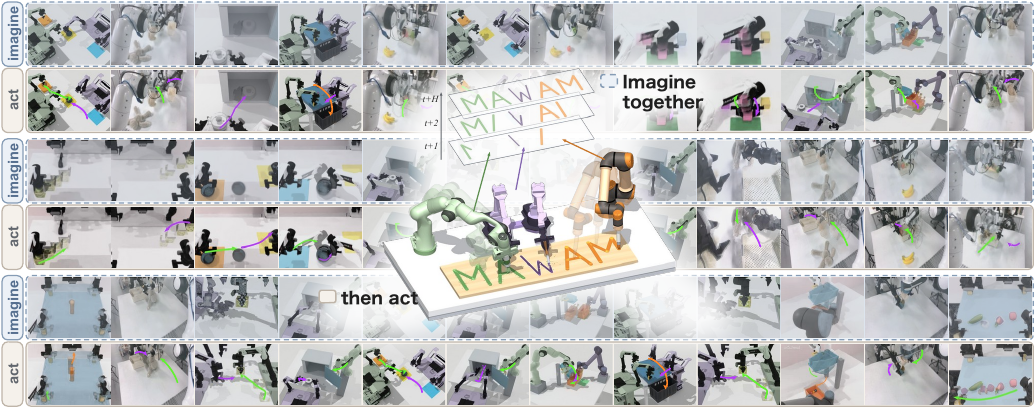}\par
\vspace{1.5pt}
\captionsetup{hypcap=false}
\captionof{figure}{\textbf{Imagine together, then act.}
\emph{Center}: a heterogeneous robot team jointly imagines one shared future over $t{+}1,\dots,t{+}H$ (cloud) before acting.
\emph{Grid}: paired \emph{imagine} and \emph{act} across simulated homo-/heterogeneous tasks and real Franka\,+\,Cobot collaboration.}
\captionsetup{hypcap=true}
\label{fig:teaser}%
\vspace{-0.2em}
\end{center}

}]
\begin{abstract}
World action models (WAMs) excel at robot manipulation by imagining future observations and decoding actions from them, yet remain confined to a single embodied agent.
In practice, manipulation often requires multiple robots to collaborate across disjoint workspaces, coupled timing, and even heterogeneous embodiments.
Such coordination demands both spatiotemporal alignment of future interactions and awareness of individual embodiment and role identities, whereas existing decentralized per-robot policies, flat centralized models, and multi-robot VLAs fail to unite both.
To bridge this gap, we present \textbf{\method}, a multi-agent world action model that jointly predicts the future observations of multiple embodiments and decodes each agent's actions from them for collaborative manipulation.
Through an agent-axis curriculum, \method expands a foundational single-agent WAM along an \emph{agent} dimension to coordinate teams under shared weights.
Within one unified causal sequence, joint self-attention spans all agents to spatiotemporally align their imagined futures, while hierarchical language, identity encodings, and embodiment masks bind each stream to its role, agent identity, and embodiment.
We further characterize collaborative manipulation, formalizing \emph{multi-agent-necessary} tasks along three axes (\emph{why}, \emph{when}, and \emph{who}) for systematic study.
Building on this formalism, we develop \textbf{\harness}, a coding-agent pipeline where mutually constraining generators and evaluators co-evolve to interactively build customizable simulation-data and evaluation packages.
Using this pipeline, we design \textbf{\bench}, following these principles across all axes with seven tasks and pairing joint success rate with individual role and collaboration scores for fine-grained analysis.
On \bench, \method reaches $82.7\%$ mean joint success, $30.7\%$ above the strongest evaluated baseline, and achieves real-time closed-loop collaboration on a real heterogeneous robot team.
\end{abstract}

\section{Introduction}
\label{sec:intro}

World action models (WAMs)~\cite{kim2026cosmospolicy,li2026lingbotva,ye2026wamzero,yuan2026fastwam,bi2026motus2,yu2026openwam} advance robot manipulation by coupling control with future visual modeling, imagining how observations evolve to ground actions in physical foresight.
Yet existing WAMs remain confined to a \emph{single embodied agent}, whereas real-world manipulation often requires multiple robots collaborating beyond one agent's limits.

These limits stem from what one agent can \textbf{observe}, where it can \textbf{reach}, what it can \textbf{actuate} and so on, often under tight temporal coupling (\eg four-arm tray transport) and across heterogeneous embodiments.
Thus, the future is no longer rolled out by one agent against a passive scene, but \emph{co-authored} across space, time, and distinct embodied agents with specialized views, kinematics, and roles.

Existing multi-robot methods fall into three categories, yet none unites spatiotemporal alignment of future interactions with awareness of individual embodiment and role identities.
First, LLM planners~\cite{mandi2023roco,liu2024coherent,chen2024emos,tan2025roboos} negotiate subtasks while delegating physical coupling to motion planners, struggling with closed-loop contacts.
Second, decentralized policies~\cite{qin2025robofactory,dong2025mimicd,doshi2026chorus} condition each robot on local views, leaving partner cues invisible and failing to align coupled futures across agents.
Third, centralized or coupled multi-robot VLAs~\cite{zhang2026mavla,zhang2026vlafabric,xu2026coopmavla} map observations reactively to actions via multi-arm prompting, peer feature exchange, or pooled views, lacking visual foresight while blurring agent, role, and embodiment identities.

Our key insight is straightforward: let a team \textbf{imagine together, then act individually} (\cref{fig:teaser}).
We present \textbf{\method}, a multi-agent world action model that jointly predicts the future observations of multiple embodiments and decodes each agent's actions from them for collaborative manipulation (\cref{sec:method}).
Through an \emph{agent-axis curriculum}, \method expands a foundational single-agent WAM along an \emph{agent} dimension from single-role trajectories to synchronized multi-robot episodes under shared weights.
Within one unified causal sequence, self-attention spans all agents to spatiotemporally align their imagined futures, while hierarchical language, Fourier identity encodings, and embodiment masks bind each stream to its role, identity, and embodiment across homogeneous and heterogeneous teams.

Studying such collaboration also exposes a benchmark gap: most manipulation suites focus on a single embodied agent (\eg RoboTwin~\cite{mu2025robotwin,chen2025robotwin2} and RoboCasa365~\cite{nasiriany2024robocasa,nasiriany2026robocasa365}), whereas multi-robot suites~\cite{mandi2023roco,qin2025robofactory} design collaborative tasks case by case: they neither verify that a task is beyond any single agent, nor systematically cover why collaboration is needed, how tightly agents couple in time, and which embodiments take part, while success-rate scoring cannot separate individual role competence from genuine collaboration.
We therefore formalize \emph{multi-agent-necessary} manipulation (\cref{sec:bench:def}) along three tested axes: \emph{why} collaboration is needed (information, spatial, or capability complementarity), \emph{when} intervals couple (asynchronous, partially synchronous, or synchronous), and \emph{who} executes each role (homogeneous or heterogeneous embodiments), paired with milestone metrics separating Role and Collaboration Scores from Success Rate.
Building on this formalism and capable coding agents, we build \textbf{\harness} (\cref{sec:bench:harness}) to co-evolve simulation generators and deterministic evaluators into customizable data-synthesis and evaluation packages, upon which we propose \textbf{\bench} (\cref{sec:bench:mabench}) with seven tasks (700 certified pairs) across all axes.
Our contributions are:
\begin{itemize}[leftmargin=*,itemsep=0pt,topsep=1pt,parsep=0pt]
  \item We formalize \emph{multi-agent-necessary} manipulation to characterize genuine multi-robot collaboration for systematic study. Grounded in this formalism, we develop \textbf{\harness}, a coding-agent pipeline where mutually constraining generators and evaluators co-evolve to interactively synthesize customizable simulation-data and evaluation packages. Using this pipeline, we construct \textbf{\bench}, covering all three axes across seven collaborative tasks and pairing joint success rate with individual role and collaboration scores for fine-grained diagnosis (\cref{sec:bench}).
  \item \textbf{\method}, a multi-agent WAM that jointly predicts spatiotemporally aligned future observations across embodiments and decodes each agent's actions by its embodiment and role identity via agent-axis expansion (\cref{sec:method}).
  \item On \bench, \method reaches $82.7\%$ mean $\SR$ ($30.7$~points above the strongest evaluated baseline), while detailed component ablations and analyses show how joint imagination and agent-axis conditioning enable cross-agent information sharing, 3-D future alignment, and timely adaptation. Finally, it achieves real-time closed-loop collaboration on a real heterogeneous robot team (\cref{sec:exp}).
\end{itemize}

\section{Related Work}
\label{sec:related}

\para{Single-agent VLAs and WAMs.}
Generalist VLAs map observations and language to actions at scale~\cite{brohan2023rt2,team2024octo,kim2024openvla,black2024pi0,black2025pi05}, while WAMs ground control in foresight by modeling future visual dynamics~\cite{du2023unipi,wu2023gr1,cheang2024gr2,hu2024vpp,li2025uva,liao2025genieenv,kim2026cosmospolicy,li2026lingbotva,ye2026wamzero,yuan2026fastwam,bi2026motus2,yu2026openwam}.
Both lines extend to diverse morphologies through cross-embodiment pretraining~\cite{oxe2023,doshi2024crossformer,wang2024hpt} and to two arms through bimanual policies~\cite{zhao2023aloha,im2025twinvla}, yet they still control a single embodied agent per rollout.
\method instead expands a single-agent WAM along an \emph{agent} axis to model a team.

\para{Multi-robot policies.}
Beyond single bodies, LLM planners negotiate subtasks and waypoints~\cite{mandi2023roco,chen2023scalablellm,liu2024coherent,chen2024emos,tan2025roboos} while leaving physical coupling to motion planners.
Learned policies output low-level actions via constrained or decentralized diffusion~\cite{qin2025robofactory,dong2025mimicd,he2025latenttom}, identity-prompted VLAs~\cite{doshi2026chorus}, multi-arm VLAs~\cite{zhang2026mavla,xu2026coopmavla}, or arm-local VLAs with learned peer interaction~\cite{zhang2026vlafabric}.
Yet all act reactively without visual foresight, whereas \method conditions actions on a \emph{jointly imagined} team future.

\para{Multi-agent world models.}
Conversely, existing multi-agent world models separate simulation from manipulation: video models predict multi-view futures \emph{given} actions~\cite{chung2026meworld,liu2026gammaworld,mo2026worldweaver,savva2026solaris,wu2026multiworld}, MARL models learn game or low-dimensional dynamics~\cite{egorov2022mambamarl,zhang2024marie,zhang2025dima,zhao2025seqwm}, COMBO plans high-level subgoals~\cite{zhang2024combo}, and driving WAMs predict trajectories~\cite{ma2026rewam,peng2026ctrlwam}.
\method instead jointly denoises multi-manipulator visual latents and outputs closed-loop actions.

\section{Defining and Building Multi-Agent Tasks}
\label{sec:bench}

\subsection{Multi-Agent Task Definition}
\label{sec:bench:def}

\para{Agents and multi-agent necessity.}
To establish when collaboration is genuinely required, consider a team of $N$ robots in a shared environment.
Each complete robot is one agent $k\in\{1,\dots,N\}$ (a dual-arm ALOHA is one agent, not two), with capability tuple $\Ecal_k=(\Ocal_k,\Wcal_k,\Ucal_k)$: what it can \emph{observe} (cameras), where it can \emph{reach} (collision-free workspace), and what it can \emph{actuate} (manipulators, DoFs, payload).
A multi-agent task is
\begin{equation}
  \tau=\big(\{\Ecal_k\}_{k=1}^{N},\,V,\,\{M_k\}_{k=1}^{N},\,\Mcollab\big),
  \label{eq:task}
\end{equation}
where $V$ is the goal predicate, $M_k$ the set of \emph{role milestones} that agent $k$ completes during its critical interval $\phi_k=[s_k,e_k]$, and $\Mcollab$ the set of \emph{collaboration milestones} that agents must satisfy jointly.
Let $\Feas(V;\Ecal)\in\{0,1\}$ indicate whether a single agent with capability $\Ecal$ can achieve $V$ alone.
We call $\tau$ \emph{multi-agent-necessary} if no agent suffices while the team succeeds:
\begin{equation}
  \forall k:\ \Feas(V;\Ecal_k)=0,\qquad \Feas\big(V;\textstyle\bigcup_{k}\Ecal_k\big)=1 .
  \label{eq:necessity}
\end{equation}
Once necessity holds, three axes structure the task.

\para{Why: complementarity.}
A single agent fails because one component of $\Ecal_k$ is insufficient.
Granting agent $k$ its partner $j$'s resources one component at a time, in order of increasing physical substitution cost ($O\prec W\prec U$), defines three relaxations:
$\Rcal^{O}(\Ecal_k)=(\Ocal_k{\cup}\Ocal_j,\Wcal_k,\Ucal_k)$,
$\Rcal^{W}(\Ecal_k)=(\Ocal_k,\Wcal_k{\cup}\Wcal_j,\Ucal_k)$ and
$\Rcal^{U}(\Ecal_k)=(\Ocal_k,\Wcal_k,\Ucal_k{\cup}\Ucal_j)$.
The \emph{minimal} relaxation restoring single-agent feasibility,
\begin{equation}
  \kappa\!=\!\min\big\{r\!\in\!(O\!\prec\! W\!\prec\! U)\!:\!\Feas\big(V;\Rcal^{r}(\Ecal_k)\big)\!=\!1\big\},
  \label{eq:kappa}
\end{equation}
names the complementarity.
\textbf{Information} ($\kappa{=}O$): task-critical cues are split across views, so no single observation resolves the state.
\textbf{Spatial} ($\kappa{=}W$): the task spans workspaces beyond any single robot, requiring transfer or handover.
\textbf{Capability} ($\kappa{=}U$): concurrent contacts, DoFs, or payload exceed any single robot.

\para{When: temporal coupling.}
Coupling is defined by whether the critical intervals of agents $j$ and $k$ must overlap.
In \textbf{asynchronous} tasks ($\phi_j\cap\phi_k=\emptyset$) agents interact only through persistent changes of the environment, which requires causal order but not simultaneity.
In \textbf{synchronous} tasks ($|\phi_j\cap\phi_k|\ge d>0$) agents act concurrently within a shared window $\phi_{jk}=\phi_j\cap\phi_k$ and must maintain an inter-agent spatiotemporal constraint at every instant of it.
\textbf{Partially synchronous} tasks lie in between: the intervals must overlap, but one agent only holds a condition quasi-statically while its partner operates, so a late partner prolongs waiting instead of breaking the constraint.

\para{Who: embodiment.}
Role assignment requires $\Feas(M_k;\Ecal_k)=1$.
A team is \textbf{homogeneous} if $\Ecal_j=\Ecal_k$, so that roles are interchangeable and distinguished only by instruction, and \textbf{heterogeneous} if $\Ecal_j\neq\Ecal_k$: role swaps may be infeasible ($\Feas(M_k;\Ecal_j)=0$), and a policy must bridge different observation and action spaces.

\para{Milestone-based metrics.}
Whereas single-agent benchmarks report only an overall stage score and success rate~\cite{mu2025robotwin,chen2025robotwin2}, we separate individual role milestones $M_k$ (\emph{who}) from inter-agent collaboration milestones $\Mcollab$ (\emph{why} and \emph{when}).
Over $E$ evaluation episodes, the completion score of a milestone set $M$ is
\begin{equation}
  S(M)=\frac{1}{E}\sum_{e=1}^{E}\frac{1}{|M|}\sum_{m\in M}\indic\big[m^{(e)}\big],
  \label{eq:scores}
\end{equation}
which gives the per-agent \textbf{Role Score} $S_k=S(M_k)$ and the \textbf{Collaboration Score} $\Scollab=S(\Mcollab)$; a collaboration milestone is credited only if the agents jointly satisfy its spatiotemporal condition within $\phi_{jk}$.
The \textbf{joint Success Rate} requires all milestones at once:
\begin{equation}
  \SR=\frac{1}{E}\sum_{e=1}^{E}\indic\Big[\bigwedge_{k}\bigwedge_{m\in M_k}\! m^{(e)}\ \wedge\!\!\bigwedge_{c\in \Mcollab}\!\! c^{(e)}\Big].
  \label{eq:sr}
\end{equation}
A policy learning local roles without coordination yields high $S_k$ yet low $\Scollab$, isolating coordination failure.

\subsection{\harness}
\label{sec:bench:harness}

\begin{figure*}[t]
  \centering
  \includegraphics[width=\textwidth]{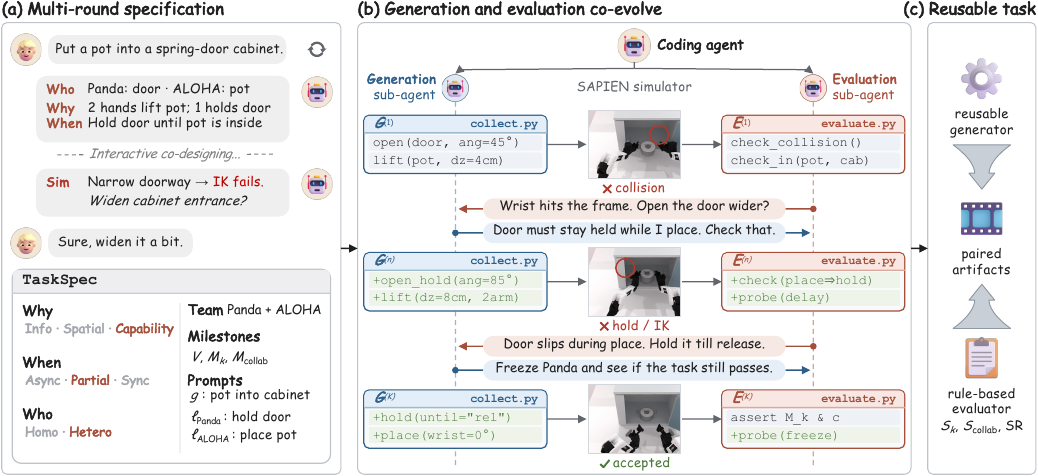}
  \caption{\textbf{\harness.}
  (a)~Through multi-round dialogue, a coding agent formalizes a task idea into a \texttt{TaskSpec} (\emph{Why}, \emph{When}, \emph{Who}), refined by simulation feedback.
  (b)~Generation and evaluation sub-agents co-audit and repair pipelines $G^{(n)}$, $E^{(n)}$ in simulation until demonstrations pass all checks.
  (c)~Converged pipelines freeze into a reusable package scored by $(S_k,\Scollab,\SR)$.}
  \label{fig:harness}
\end{figure*}

Turning this formal definition into working tasks normally demands heavy manual engineering of multi-robot planners, timing windows, and matching evaluation functions, where any mismatch yields uncertified demonstrations or flawed scores.
To automate this process, \harness (\cref{fig:harness}) is a coding-agent framework that converts a task idea into a verified, reusable package in three stages.

\para{(1) Specification via Why--When--Who.}
Starting from a high-level idea (\eg ``put a pot into a spring-door cabinet''), the coding agent clarifies the task with the user over several rounds (\cref{fig:harness}a).
\emph{Who}: it fixes the team $\{\Ecal_k\}$, decomposes role milestones $M_k$ and drafts a global team instruction $g$ and per-robot role instructions $\ell^k$.
\emph{Why and When}: it identifies why a single agent fails ($\kappa$) and how the critical intervals couple, which defines $\Mcollab$ (\eg the door stays held while the pot is placed).
When simulation reveals a physical bottleneck (\eg a doorway too narrow for IK), it returns a diagnosis with a proposed fix.
The dialogue yields a structured \texttt{TaskSpec} with labels, team, prompts and milestones.

\para{(2) Co-evolution of generation and evaluation.}
From the \texttt{TaskSpec}, the harness spawns two sub-agents that refine each other over rounds $n=1,\dots,K$ in a RoboTwin-derived simulator~\cite{mu2025robotwin,chen2025robotwin2} (\cref{fig:harness}b).
The \emph{generation pipeline} $G^{(n)}$ synthesizes scene layouts and multi-robot expert planners and accepts a trajectory only if it passes physical checks (collision-free execution, continuous IK, stable contacts) and the predicates $c\in\Mcollab$.
The \emph{evaluation pipeline} $E^{(n)}$ compiles the \texttt{TaskSpec} into deterministic, LLM-free detectors over simulator state: it tracks $M_k$ and $\Mcollab$ to compute $(S_k,\Scollab,\SR)$ and runs \emph{necessity probes}, \eg delaying or freezing one partner, that verify the declared labels.
Physical failures and violated hold windows make the generator repair its planning and timing; false rejections and violations missed under the probes make the evaluator extend its rules.
Sharing predicates across acceptance and scoring makes both pipelines converge to one task semantics.

\para{(3) Reusable task package.}
The harness freezes a self-contained package (\cref{fig:harness}c): a \emph{generator} for demonstration synthesis, \emph{paired multi-agent artifacts} (synchronized multi-view RGB, native actions, hierarchical prompts), and a deterministic \emph{evaluator} scoring policies with unified metrics.

\subsection{\bench}
\label{sec:bench:mabench}

\begin{figure*}[t]
  \centering
  \includegraphics[width=\textwidth]{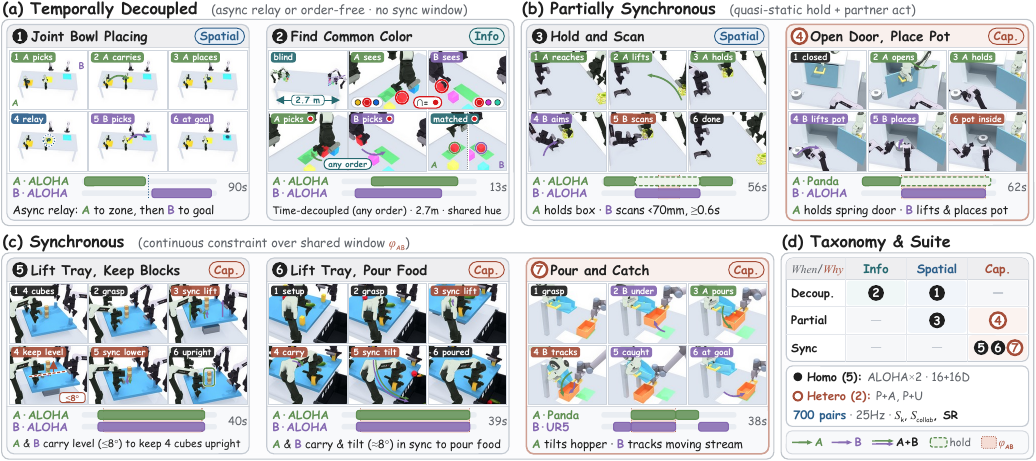}
  \caption{\textbf{\bench.}
  (a--c)~Studio renders, trajectories, critical-interval timelines and rules of the seven tasks.
  (d)~\emph{When}$\times$\emph{Why} coverage and team specs (hollow: heterogeneous).}
  \label{fig:bench}
\end{figure*}

Following \cref{sec:bench:def} and using the \harness, we build \bench, a suite of seven collaborative tasks (\cref{fig:bench}).
\textbf{Asynchronous}: in \emph{Joint bowl placing} (spatial) one ALOHA moves a bowl into a shared zone and the other carries it to a goal outside the first robot's reach; in \emph{Find common color} (information) two ALOHAs $2.7$\,m apart see disjoint sets of blocks sharing exactly one color, which each must pick on its own side.
\textbf{Partially synchronous}: in \emph{Hold and scan} (spatial) one ALOHA presents a box while the other aligns a scanner with it; in \emph{Open door, place pot} (capability) a single-arm Panda holds a spring-loaded cabinet door open while an ALOHA carries a pot inside with both arms.
\textbf{Synchronous} (capability): two ALOHAs carry a tray with four arms without toppling the stack on it (\emph{Lift tray, keep blocks}) or tilt it in unison to pour its contents into a bag (\emph{Lift tray, pour food}); in \emph{Pour and catch} a Panda tilts a hopper while a UR5 tracks the moving stream with a cup.

Across the suite, five tasks pair homogeneous ALOHAs and two pair heterogeneous embodiments; in every task, each robot records three RGB views (head and two wrists, with missing single-arm wrist views masked), proprioception, and native end-effector actions at 25\,Hz.
In total, \bench provides 700 accepted demonstration pairs (100 per task), with every task's axis labels certified by its evaluator's necessity probes (details in \cref{sec:supp:def,sec:supp:bench}).
\bench spans all temporal regimes, complementarity types, and team compositions in closed loop (\cref{tab:bench_compare}, \cref{sec:supp:bench}).

\section{\method}
\label{sec:method}

\begin{figure*}[t]
  \centering
  \includegraphics[width=\textwidth]{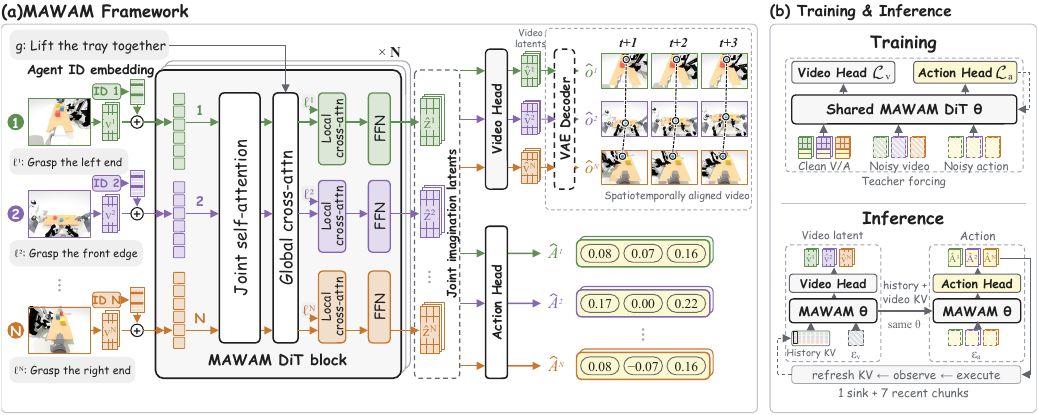}
  \caption{\textbf{Overview of \method.}
  (a)~Each robot provides its views and role $\ell^k$, with fixed identity and role encodings added once at the input.
  Every shared DiT block runs joint self-attention across \emph{all} robots, gated global cross-attention to $g$, and local cross-attention to $\ell^k$, producing a spatiotemporally aligned \emph{joint imagination latent} $\hat z^{1:N}$ from which the action head decodes $\hat A^{k}$.
  (b)~Training supervises video and action jointly; inference denoises joint video latents before actions, then grounds the sliding joint KV cache with real observations.}
  \label{fig:method}
\end{figure*}

To meet the collaboration demands of \cref{sec:bench}, \method unifies the team in one generative network that jointly imagines all robots' future views and decodes their actions (\cref{fig:method}).

\subsection{Problem Formulation}
\label{sec:method:problem}

At control step $t$ (25\,Hz), robot $k$ with embodiment $e_k$ (Panda, UR5 or dual-arm ALOHA~\cite{zhao2023aloha}) receives three RGB views $o_t^k$ and issues an action $a_t^k$.
The team is given a global instruction $g$ and every robot a role instruction $\ell^k$.
Actions are end-effector targets (8-D for single-arm and 16-D for dual-arm robots) written into a shared $30$-D slot layout with a binary embodiment mask $m_{e_k}$ over active channels.
Video is subsampled to 6.25\,Hz and encoded by a causal video VAE~\cite{wan2025wan}, giving one latent frame $z_f^k$ per 16 control steps, paired with its action block $A_f^k\in\mathbb{R}^{16\times 30}$.
Let $\Hcal_f=\{(z^{k}_{\le f},A^{k}_{\le f})\}_{k=1}^{N}$ be the team history.
Predicting $C$-frame chunks, \method models the \emph{joint} conditional of chunk $c=(f,f{+}C]$ as
\begin{equation}
  \underbrace{p_\theta\big(z_c^{1:N}\,\big|\,\Hcal_f,\ell^{1:N},g\big)}_{\text{imagine together}}\;
  \underbrace{p_\theta\big(A_c^{1:N}\,\big|\,z_c^{1:N},\Hcal_f,\ell^{1:N},g\big)}_{\text{then act}} .
  \label{eq:joint}
\end{equation}
Whereas per-robot deployment reduces \cref{eq:joint} to a product of marginals $\prod_k p(z_c^k|\Hcal_f^k)\,p(A_c^k|z_c^k,\Hcal_f^k)$ whose imagined futures need not agree, \cref{eq:joint} conditions robot $k$'s future on both robot $j$'s history \emph{and} its concurrently imagined future.

\subsection{Architecture}
\label{sec:method:arch}

\para{Backbone.}
We build on \lingbot~\cite{li2026lingbotva}, an autoregressive WAM on a video diffusion transformer~\cite{peebles2022dit,wan2025wan} with 30 blocks and about 5\,B parameters.
Each block applies self-attention with 3-D rotary position encoding, cross-attention to a frozen T5 text encoder~\cite{raffel2020t5} and an FFN; separate heads predict video and action velocities.
Both modalities are trained with flow matching~\cite{lipman2022flow,liu2022rectified}: for clean $x$, noise $\epsilon$ and level $\sigma$, the model receives $x^\sigma=(1{-}\sigma)x+\sigma\epsilon$ and regresses $u=\epsilon-x$, with an independent $\sigma_f$ per latent frame as in diffusion forcing~\cite{chen2024diffforcing}.

\para{Causal structure.}
Training uses teacher forcing with two copies of every frame, a \emph{noisy} prediction target and a \emph{clean} history.
A stage index $s=2\lfloor f/C\rfloor+\mu$, with $\mu{=}0$ for video and $\mu{=}1$ for action, orders video before action within a chunk.
A query token $q$ may attend to a key token $v$ iff $|s_q-s_v|\le W$ and
\begin{equation}
\begin{aligned}
  &\text{clean}\!\to\!\text{clean}: s_v\le s_q, &&\text{noisy}\!\to\!\text{clean}: s_v< s_q,\\
  &\text{noisy}\!\to\!\text{noisy}: s_v= s_q, &&\text{clean}\!\to\!\text{noisy}: \text{never}.
\end{aligned}
\label{eq:mask}
\end{equation}
Noisy actions of chunk $c$ thus see the clean video of chunk $c$: actions are an inverse-dynamics readout of the imagination.

\para{The agent axis: joint causal attention.}
\method adds robots as a token axis and shares all parameters $\theta$ across them.
Per-robot sequences are concatenated into length $N{\cdot}L$, projected by shared query/key/value maps, and processed by joint self-attention under \cref{eq:mask}.
Crucially, \emph{the robot index never gates an attention edge}, directly realizing \cref{eq:joint}:
(i)~robot $k$'s noisy video attends to all robots' clean histories and same-stage noisy videos, denoising team futures \emph{jointly} into one imagination;
(ii)~robot $k$'s noisy actions attend to \emph{all} robots' clean videos in the chunk, acting on partners' imagined futures alongside its own.
At $N{=}1$ this recovers the backbone, with parameter count independent of $N$.

\para{Hierarchical language.}
We keep the backbone's cross-attention as a \emph{local} pathway where robot $k$ attends only to its role tokens $L_k$ (from $\ell^k$), and add a gated \emph{global} cross-attention $\CA^{g}_b$ in every block $b$ exposing goal tokens $G$ (from $g$) to all robots.
From post-self-attention states $\tilde h$,
\begin{equation}
  h'\!=\!\tilde h\!+\!\gamma_b\CA^{g}_b(\LN^{g}_b(\tilde h),\!G),\;
  h''_k\!=\!h'_k\!+\!\CA^{\ell}_b(\LN_b(h'_k),\!L_k),
\label{eq:hier}
\end{equation}
followed by the FFN; $\CA^{g}_b$ is initialized from the local module with gate $\gamma_b{=}10^{-3}$ to start at the single-agent model.

\para{Identity and role encoding.}
Since all robots share weights and position codes, robot $k$ receives a fixed physical ID $n_k\in\{0,1,\dots\}$ via unit-RMS Fourier features $\phi(n_k)$ and a normalized mean role vector $r_k$ of $L_k$, added once after input projection: $h^{(0)}_k\leftarrow h^{(0)}_k+\beta(r_k+\phi(n_k))$ ($\beta{=}0.01$).
Without learned ID tables or robot-specific branches, heterogeneous embodiments are expressed solely via identity, role text, and view/action masks.
Each module targets one axis of \cref{sec:bench:def}: partner attention resolves \emph{information} tasks, joint denoising aligns \emph{synchronous} tasks, hierarchical language divides labor, and identities with masks bridge \emph{heterogeneous} teams.

\begin{table*}[t]
\centering
\definecolor{t2bestbg}{HTML}{D6E6F6}%
\definecolor{t2gaincol}{HTML}{137333}%
\definecolor{t2submet}{HTML}{525C68}%
\definecolor{t2zerocol}{HTML}{9AA4B0}%
\def\tIms#1{{\color{t2submet}#1}}%
\def\tImz{{\color{t2zerocol}0.0}}%
\def\tIbest#1{\cellcolor{t2bestbg}\textbf{#1}}%
\def\tIsecb#1{\underline{#1}}%
\def\tIoa#1#2{\cellcolor{t2bestbg}\textbf{#1}\,{\fontsize{6.5}{7.5}\selectfont\color{t2gaincol}\textbf{(+#2\%)}}}%
\def\tIsc{$S_{\!\mathrm{collab}}$}%
\caption{\textbf{Main results on \bench} ($100$ independent trajectory trials per task): Role Score $S_k$ (mean over agents), Collaboration Score $\Scollab$, and joint Success Rate $\SR$ (\%).
\textbf{Bold} (blue) and \underline{underline} mark the best and second-best $\SR$ (95\% Wilson CIs in \cref{sec:supp:results}); parenthesized values on the average row denote \textbf{\method}'s gains over the strongest baseline.}
\label{tab:main}
\vspace{-0.55em}
{\footnotesize
\setlength{\tabcolsep}{1.75pt}%
\renewcommand{\arraystretch}{1.06}%
\resizebox{\textwidth}{!}{%
\begin{tabular}{@{}l@{\hspace{5pt}} ccc@{\hspace{5pt}} ccc@{\hspace{6pt}} ccc@{\hspace{5pt}} ccc@{\hspace{6pt}} ccc@{\hspace{5pt}} ccc@{\hspace{6pt}} ccc@{\hspace{2.2pt}}}
\toprule
 & \multicolumn{6}{c@{\hspace{6pt}}}{Per-robot} & \multicolumn{6}{c@{\hspace{6pt}}}{All-view} & \multicolumn{6}{c@{\hspace{6pt}}}{Multi-robot VLA} & \multicolumn{3}{c@{\hspace{2.2pt}}}{\textbf{Ours}} \\
\cmidrule(r{6pt}){2-7}\cmidrule(r{6pt}){8-13}\cmidrule(r{6pt}){14-19}\cmidrule{20-22}
 & \multicolumn{3}{c@{\hspace{5pt}}}{\pizf~\cite{black2025pi05}} & \multicolumn{3}{c@{\hspace{6pt}}}{\lingbot~\cite{li2026lingbotva}} & \multicolumn{3}{c@{\hspace{5pt}}}{\pizf (joint)} & \multicolumn{3}{c@{\hspace{6pt}}}{\lingbot (joint)} & \multicolumn{3}{c@{\hspace{5pt}}}{MA-VLA~\cite{zhang2026mavla}} & \multicolumn{3}{c@{\hspace{6pt}}}{VLA-Fabric~\cite{zhang2026vlafabric}} & \multicolumn{3}{c@{\hspace{2.2pt}}}{\textbf{\method}} \\
\cmidrule(r{5pt}){2-4}\cmidrule(r{6pt}){5-7}\cmidrule(r{5pt}){8-10}\cmidrule(r{6pt}){11-13}\cmidrule(r{5pt}){14-16}\cmidrule(r{6pt}){17-19}\cmidrule{20-22}
Task & \tIms{$S_k$} & \tIms{\tIsc} & $\SR$ & \tIms{$S_k$} & \tIms{\tIsc} & $\SR$ & \tIms{$S_k$} & \tIms{\tIsc} & $\SR$ & \tIms{$S_k$} & \tIms{\tIsc} & $\SR$ & \tIms{$S_k$} & \tIms{\tIsc} & $\SR$ & \tIms{$S_k$} & \tIms{\tIsc} & $\SR$ & \tIms{$S_k$} & \tIms{\tIsc} & $\SR$ \\
\midrule
Joint bowl placing     & \tIms{33.0} & \tIms{4.0}  & \tImz & \tIms{35.5} & \tIms{6.5}  & 5.0          & \tIms{24.8} & \tImz      & \tImz & \tIms{38.0}  & \tIms{16.0} & 5.0          & \tIms{31.3} & \tIms{3.0}  & \tImz        & \tIms{40.8} & \tIms{13.5} & \tIsecb{9.0} & \tIms{83.3}  & \tIms{77.0} & \tIbest{66.0} \\
Find common color      & \tIms{56.8} & \tIms{8.5}  & 6.0   & \tIms{98.8} & \tIms{12.5} & 11.0         & \tIms{0.5}  & \tImz      & \tImz & \tIms{98.0}  & \tIms{30.5} & 28.0         & \tIms{84.0} & \tIms{37.5} & \tIsecb{35.0}& \tIms{87.0} & \tIms{35.5} & 33.0         & \tIms{100.0} & \tIms{66.5} & \tIbest{63.0} \\
Hold and scan          & \tIms{36.3} & \tIms{30.5} & 26.0  & \tIms{93.0} & \tIms{90.5} & \tIsecb{82.0}& \tImz       & \tImz      & \tImz & \tIms{24.3}  & \tImz       & \tImz        & \tIms{20.0} & \tIms{2.0}  & \tImz        & \tIms{60.5} & \tIms{47.5} & 42.0         & \tIms{98.0}  & \tIms{97.0} & \tIbest{95.0} \\
Open door, place pot   & \tIms{42.8} & \tIms{29.0} & 21.0  & \tIms{95.3} & \tIms{90.5} & \tIsecb{81.0}& \tImz       & \tImz      & \tImz & \tIms{25.5}  & \tIms{0.5}  & \tImz        & \tIms{56.5} & \tIms{40.0} & 31.0         & \tIms{23.8} & \tIms{4.5}  & \tImz        & \tIms{97.3}  & \tIms{94.5} & \tIbest{88.0} \\
Lift tray, keep blocks & \tIms{7.5}  & \tIms{2.3}  & \tImz & \tIms{66.5} & \tIms{59.0} & \tIsecb{47.0}& \tImz       & \tImz      & \tImz & \tImz        & \tImz       & \tImz        & \tIms{7.5}  & \tIms{2.3}  & \tImz        & \tIms{12.0} & \tIms{3.7}  & \tImz        & \tIms{95.5}  & \tIms{93.7} & \tIbest{93.0} \\
Lift tray, pour food   & \tIms{21.5} & \tIms{11.5} & \tImz & \tIms{85.3} & \tIms{67.0} & 41.0         & \tImz       & \tImz      & \tImz & \tIms{100.0} & \tIms{81.5} & \tIsecb{63.0}& \tIms{85.0} & \tIms{61.0} & 38.0         & \tIms{86.8} & \tIms{61.0} & 36.0         & \tIms{100.0} & \tIms{95.5} & \tIbest{91.0} \\
Pour and catch         & \tIms{22.3} & \tIms{20.0} & 17.0  & \tIms{98.8} & \tIms{98.0} & \tIbest{97.0}& \tIms{5.5}  & \tIms{0.5} & \tImz & \tIms{0.8}   & \tImz       & \tImz        & \tIms{7.3}  & \tIms{2.0}  & \tImz        & \tIms{7.8}  & \tIms{2.0}  & \tImz        & \tIms{85.8}  & \tIms{86.0} & \tIsecb{83.0} \\
\midrule
\textbf{Average}       & 31.4 & 15.1 & 10.0 & \tIsecb{81.9} & \tIsecb{60.6} & \tIsecb{52.0} & 4.4 & 0.1 & \tImz & 40.9 & 18.4 & 13.7 & 41.6 & 21.1 & 14.9 & 45.5 & 24.0 & 17.1 & \tIoa{94.3}{12.4} & \tIoa{87.2}{26.6} & \tIoa{82.7}{30.7} \\
\bottomrule
\end{tabular}}}
\end{table*}

\subsection{Agent-Axis Curriculum}
\label{sec:method:train}

Because $N{=}1$ recovers the single-agent backbone under shared weights, both stages train the \emph{same} network and differ only in agent-axis length.
\textbf{Stage~1 ($N{=}1$):} demonstrations are split into per-robot trajectories with own views, actions, and role $\ell^k$ (heterogeneous robots mixed 1:1).
\textbf{Stage~2 ($N{>}1$):} synchronized episodes form joint sequences sharing per-frame noise levels $\sigma_f$ with independent noise draws:
\begin{equation}
  \Lcal\!=\!\tfrac{1}{N}\!\sum_{k=1}^{N}\!\Big(\Lcal_{\mathrm{vid}}^{k}\!+\!\tfrac{1}{F}\!\sum_{f}\!w(\sigma_f)\frac{\|m_{e_k}\!\odot\!(\hat u^{k}_{f}\!-\!u^{k}_{f})\|_2^2}{16\,\|m_{e_k}\|_1}\Big),
  \label{eq:loss}
\end{equation}
combining masked action and view-masked video ($\Lcal_{\mathrm{vid}}^{k}$) losses.
All local and global prompts drop jointly with probability $0.1$ for CFG~\cite{ho2022cfg}; all weights are fine-tuned with a $10\times$ learning rate on $\gamma_b$ (details in \cref{sec:supp:model}).

\subsection{Closed-Loop Inference}
\label{sec:method:infer}

At deployment, one \method instance controls the team in closed loop via a sliding \emph{joint KV cache} ($W{=}64$ latent frames) storing all robots' clean history (\cref{fig:method}b).
Every chunk ($C{=}2$ latent frames, 32 steps or 1.28\,s) runs four steps:
\textbf{(1)~Imagine jointly:} denoise all robots' future video latents together;
\textbf{(2)~Act:} cache imagined latents as clean tokens and jointly denoise all robots' action blocks, re-applying $m_{e_k}$ each step;
\textbf{(3)~Execute:} map slots to native end-effector commands;
\textbf{(4)~Ground:} overwrite the imagined chunk in the cache with encoded real observations and executed actions.
Benchmark evaluation uses 25/50 video/action steps; an accelerated 4/8-step real-time setup with fused QKV projections, compiled RoPE, and cached text KVs runs in $0.72$\,s per chunk on one GPU for pause-free 30\,Hz control (details in \cref{sec:supp:model,sec:supp:real}).

\section{Experiments}
\label{sec:exp}

\subsection{Setup}
\label{sec:exp:setup}

We evaluate seven policies across all seven \bench tasks ($100$ independent trajectory trials per task under synchronous evaluation) on Role Score $S_k$, Collaboration Score $\Scollab$, and joint Success Rate $\SR$ (\cref{sec:bench:def}):
(i)~\emph{Per-robot} (\pizf~\cite{black2025pi05}, \lingbot~\cite{li2026lingbotva}/Stage~1), running independent instances on each robot's views, history, and role $\ell^k$;
(ii)~\emph{All-view} (\pizf~(joint), \lingbot~(joint)), pooling all views and prompts into one flat stream---\lingbot~(joint) thus serves as a backbone-matched WAM control lacking our agent axis;
(iii)~\emph{Multi-robot VLAs}, covering centralized decomposed-prompt \textbf{MA-VLA}~\cite{zhang2026mavla} and peer-coupled arm-local \textbf{VLA-Fabric}~\cite{zhang2026vlafabric}; and our
(iv)~\textbf{\method}.
All methods share identical demonstrations, action interfaces, and evaluators (details in \cref{sec:supp:model,sec:supp:baselines}).

\subsection{Main Results}
\label{sec:exp:main}

\method leads \bench on every metric (\cref{tab:main}), reaching \textbf{94.3\%} $S_k$ ($+12.4$~points over the best baseline), \textbf{87.2\%} $\Scollab$ ($+26.6$), and \textbf{82.7\%} $\SR$ ($+30.7$).
It ranks first on \textbf{6 of 7} tasks, with gains widening from local roles ($+12.4$) to coordination ($+26.6$) and joint $\SR$ ($+30.7$).

\para{Quantitative analysis.}
\emph{Decentralized policies} preserve single-robot priors ($81.9\%$ $S_k$ for \lingbot) yet drop sharply in $\Scollab$ ($60.6\%$), acting blindly on partner cues (\emph{Common color}: $11.0\%$ $\SR$) or drifting out of phase on 4-arm transport ($41.0\%$--$47.0\%$).
\emph{Multi-robot VLAs} partially share static cues (\emph{Common color}: $33.0\%$--$35.0\%$), yet reactive coupling without visual foresight fails to align multi-body trajectories ($14.9\%$ and $17.1\%$ mean $\SR$).
Conversely, \emph{flat all-view pooling} entangles agent and embodiment identities, halving \lingbot's $S_k$ ($40.9\%$, $13.7\%$ $\SR$) and collapsing to $0.0\%$ on both heterogeneous tasks, whereas \method sustains $81.6\%$ (homogeneous) and $85.5\%$ (heterogeneous) $\SR$ (\cref{tab:axes}).

\begin{figure*}[t]
  \centering
  \includegraphics[width=0.97\textwidth]{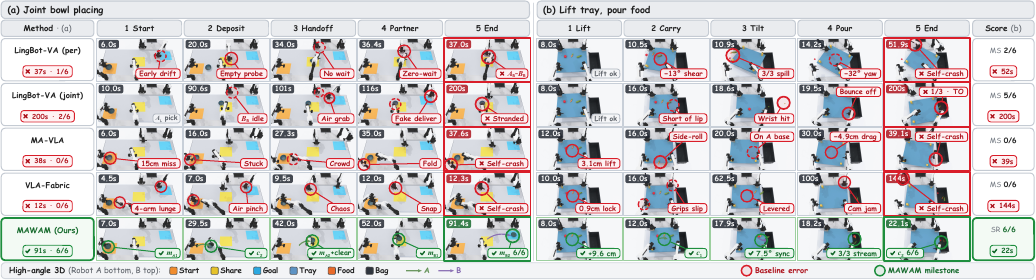}
  \vspace{-0.55em}
  \caption{\textbf{Qualitative comparison on \bench} for (a)~\emph{Joint bowl placing} (async) and (b)~\emph{Lift tray, pour food} (sync $SE(3)$): red circles mark baseline collisions, shear, and role blur; green circles verify \method's $6/6$ milestones ($8.5$\,s clearance on (a), locked pour on (b)).}
  \label{fig:qual_dual}
  \vspace{-0.65em}
\end{figure*}

\para{Qualitative analysis.}
\cref{fig:qual_dual} mirrors these metric gaps physically: \emph{per-robot} policies retain local motions yet lack shared foresight---probing an empty zone to collide ($37.0$\,s) on (a) and shearing the 4-arm chain ($-32^\circ$ spill) on (b), matching their high $S_k$ but low $\Scollab$; \emph{joint} flat pooling entangles arm ownership and precision---leaving $B_R$ idle while $B_L$ fake-delivers air on (a) and bouncing off the rim on (b), echoing its halved $S_k$---while reactive VLAs self-crash.
By contrast, \method jointly imagines aligned team futures under preserved role and embodiment identities, securing an $8.5$\,s clearance window on (a) and locked 4-arm $SE(3)$ pouring on (b) ($6/6$ milestones).

\subsection{Ablations and Analysis}
\label{sec:exp:ablation}
\label{sec:exp:analysis}

\begin{table}[t]
\centering\footnotesize\setlength{\tabcolsep}{2.0pt}
\definecolor{t3bestbg}{HTML}{D6E6F6}%
\def\tIIIbest#1{\cellcolor{t3bestbg}\textbf{#1}}%
\caption{\textbf{Ablations} on \emph{Joint bowl placing}, \emph{Open door, place pot}, and \emph{Lift tray, pour food} (\%); \textbf{bold} (blue) marks the best.}
\label{tab:ablation}
\vspace{-0.5em}
{\renewcommand{\arraystretch}{0.95}%
\begin{tabular}{@{}l cc cc cc@{}}
\toprule
& \multicolumn{2}{c}{\makecell[b]{Joint bowl\\[-1pt]placing}} & \multicolumn{2}{c}{\makecell[b]{Open door,\\[-1pt]place pot}} & \multicolumn{2}{c@{}}{\makecell[b]{Lift tray,\\[-1pt]pour food}} \\
\cmidrule(lr){2-3}\cmidrule(lr){4-5}\cmidrule(l){6-7}
Variant & $\Scollab$ & $\SR$ & $\Scollab$ & $\SR$ & $\Scollab$ & $\SR$ \\
\midrule
\textbf{Full \method}                    & \tIIIbest{77.0} & \tIIIbest{66.0} & 94.5            & 88.0            & \tIIIbest{95.5} & \tIIIbest{91.0} \\
(a) w/o joint self-attention             &  0.0            &  0.0            &  0.0            &  0.0            & 50.0            &  0.0            \\
\midrule
(b) w/o global instruction $g$           & 40.0            & 30.0            & \tIIIbest{95.0} & \tIIIbest{90.0} & 95.0            & 90.0            \\
(c) w/o local roles $\ell^k$             &  0.0            &  0.0            &  0.0            &  0.0            &  0.0            &  0.0            \\
(d) w/o identity/role encoding           & 32.5            & 20.0            & 17.5            &  0.0            & 27.5            &  0.0            \\
\bottomrule
\end{tabular}}
\end{table}

\para{Joint spatiotemporal imagination.}
As shown in \cref{tab:ablation}(a), removing cross-agent joint self-attention over the upcoming chunk so that each robot denoises its future independently from shared history $\Hcal_f$ causes the policy to fail ($0.0\%$ $\SR$) across all three temporal regimes.
Specifically, $\Scollab$ drops to $0.0\%$ on asynchronous \emph{Joint bowl placing} ($77.0\%\!\to\!0.0\%$) and partially synchronous \emph{Open door, place pot} ($94.5\%\!\to\!0.0\%$) because neither robot anticipates zone clearance or door holding, while on synchronous \emph{Lift tray, pour food}, symmetric starts lift the tray ($50.0\%$ $\Scollab$) before unaligned tilt predictions drift out of phase and spill ($91.0\%\!\to\!0.0\%$ $\SR$).

\para{Hierarchical language and identity encoding.}
Under weight-shared agent streams, local roles and identities are foundational: removing local roles $\ell^k$ (\cref{tab:ablation}c) causes complete failure across all three tasks, and retaining $\ell^k$ without identity/role encodings (\cref{tab:ablation}d) similarly collapses performance, showing that the agent axis is vital to multi-robot collaboration by binding each role and visual-action stream to its corresponding embodiment branch.
Meanwhile, global instruction $g$ (\cref{tab:ablation}b) proves indispensable for unifying decoupled subtasks: in \emph{Joint bowl placing}, agents act without physical coupling, so removing $g$ loses shared team intent and halves $\SR$ ($66.0\%\!\to\!30.0\%$, $\Scollab$: $77.0\%\!\to\!40.0\%$), whereas under physical coupling via distinct bodies or co-grasps (\emph{Open door, place pot}, \emph{Lift tray, pour food}), the gated global pathway stays non-intrusive.

\begin{figure}[t]
  \centering
  \includegraphics[width=0.96\linewidth]{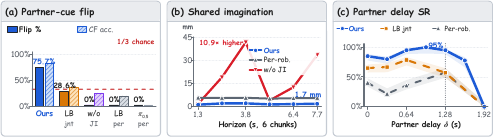}
  \vspace{-0.45em}
  \caption{\textbf{Diagnostic probes}: (a)~partner-cue flip, (b)~3-D imagination disagreement, and (c)~partner delay $\delta$ (\emph{Pour food}).}
  \label{fig:analysis}
  \vspace{-0.65em}
\end{figure}

\para{Does \method leverage other agents' information?}
In \emph{Find common color}, each robot sees only its three blocks to pick the color shared with its unseen partner; keeping own initial views fixed, we recolor only the partner's blocks to shift the shared color ($50$ trials).
In \cref{fig:analysis}a, \method flips to the new target in $75.7\%$ of solved scenes ($82.0\%$ accuracy vs.\ $33.3\%$ chance), vs.\ $28.6\%$ ($36.0\%$) for \lingbot~(joint) and $0.0\%$ w/o joint imagination, confirming that \method genuinely leverages information from other agents.

\para{Are predicted futures spatiotemporally aligned?}
On \emph{Lift tray, keep blocks}, we decode $7.68$\,s imagined videos ($20$ scenes) and probe 3-D tray pose gaps (\cref{sec:supp:consistency}).
In \cref{fig:analysis}b, \method holds disagreement at $1.74$\,mm ($3.2\times$ below per-robot \lingbot) vs.\ $18.74$\,mm ($10.9\times$) w/o joint imagination, spatiotemporally aligning 3-D futures.

\para{Can \method respond timely to partner states?}
On \emph{Lift tray, pour food} ($\le{\sim}8^\circ$ tilt), we freeze Robot~B at $t{=}0$ for $\delta\in[0,1.92]$\,s while Robot~A runs closed-loop ($20$ scenes).
In \cref{fig:analysis}c, \method keeps $80\%$--$100\%$ $\SR$ up to $1.28$\,s delay ($95.0\%$ vs.\ $57.1\%$ baselines) until the $1.92$\,s tilt bound, timely responding in real-time closed-loop collaboration.

\subsection{Real-Robot Experiments}
\label{sec:exp:real}

\begin{figure}[!t]
  \centering
  \includegraphics[width=0.96\linewidth]{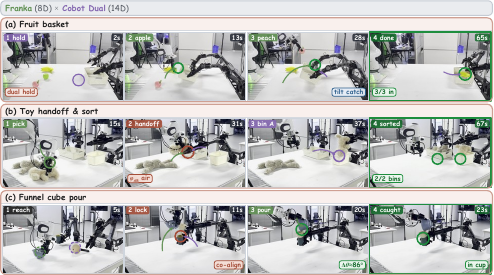}
  \vspace{-0.4em}
  \caption{\textbf{Real-robot tasks} on Franka\,+\,Cobot: (a)~\emph{Fruit basket}, (b)~\emph{Toy handoff \& sort}, and (c)~\emph{Funnel cube pour}.}
  \label{fig:real}
  \vspace{-0.65em}
\end{figure}

We evaluate \method on three physical tasks aligned with \bench, \emph{Fruit basket} (partial-sync, spatial), \emph{Toy handoff \& sort} (sync $\varphi_{AB}$, spatial), and \emph{Funnel cube pour} (sync $SE(3)$, capability), against per-robot and flat-joint \lingbot on a heterogeneous Franka ($8$-D) and dual-arm Cobot ($14$-D) cell ($3$ views/robot, $30$\,Hz; \cref{fig:real}).
Each policy is trained on $80$ teleoperated pairs per task and evaluated over $20$ trials (\cref{tab:real}; \cref{sec:supp:real}), using $4$ video and $8$ action denoising steps paired with asynchronous parallel decoding for real-time closed-loop control.

\begin{table}[!ht]
\vspace{-0.3em}
\centering\footnotesize\setlength{\tabcolsep}{3.8pt}
\definecolor{t4bestbg}{HTML}{D6E6F6}%
\def\tIVbest#1{\cellcolor{t4bestbg}\textbf{#1}}%
\caption{\textbf{Real-robot evaluation results.} Joint success rates ($\SR$, \%) on Franka\,+\,Cobot ($20$ trials/task; \textbf{bold}: best).}
\label{tab:real}
\vspace{2pt}
{\renewcommand{\arraystretch}{0.96}%
\begin{tabular}{@{}l ccc@{}}
\toprule
Task & \makecell[b]{\lingbot\\[-1pt](per-robot)} & \makecell[b]{\lingbot\\[-1pt](joint)} & \makecell[b]{\textbf{\method}\\[-1pt]\textbf{(ours)}} \\
\midrule
Fruit basket        & 15.0 & 0.0 & \tIVbest{75.0} \\
Toy handoff \& sort &  0.0 & 0.0 & \tIVbest{50.0} \\
Funnel cube pour    & 25.0 & 0.0 & \tIVbest{65.0} \\
\midrule
\textbf{Average}    & 13.3 & 0.0 & \tIVbest{63.3} \\
\bottomrule
\end{tabular}}
\vspace{-0.4em}
\end{table}

\method attains the highest success rate on all three tasks ($63.3\%$ mean vs.\ $13.3\%$ per-robot and $0.0\%$ joint).
Consistent with simulation (\cref{sec:exp:main}), per-robot \lingbot executes local motions but lacks shared foresight to synchronize mid-air handoffs ($0.0\%$ on \emph{Toy handoff \& sort}), while \lingbot~(joint) entangles both embodiments in one flat stream and completes no trial, mirroring its heterogeneous simulation collapse.
By contrast, \method coordinates both robots reliably on hardware, with remaining failures mainly caused by imprecise single-robot grasping or alignment rather than coordination errors.

\vspace{-0.55em}
\section{Conclusion}
\label{sec:conclusion}

In this work, we present \method, a multi-agent world action model that jointly predicts the future observations of multiple embodiments and decodes each agent's actions from them for collaborative manipulation.
To systematically study multi-robot collaboration, we formalize \emph{multi-agent-necessary} tasks along three axes (\emph{why}, \emph{when}, and \emph{who}), build \harness to co-evolve customizable simulation-data and evaluation packages via coding agents, and establish \bench across seven collaborative tasks with fine-grained role and collaboration metrics.
Extensive experiments qualitatively and quantitatively validate \method's effectiveness over prior baselines, while component ablations and diagnostic probes verify how joint imagination and agent-axis conditioning drive cross-agent information sharing, 3-D future alignment, and timely partner adaptation; finally, real-time closed-loop deployment on a heterogeneous physical robot team demonstrates reliable real-world collaboration.

\clearpage
{
    \small
    \bibliographystyle{ieeenat_fullname}
    \bibliography{references}
}

\clearpage
\twocolumn[{%
  \centering
  \vspace*{0.2em}
  {\Large\textbf{Appendix}\par}
  \vspace{1.0em}
}]
\appendix
\renewcommand{\dbltopfraction}{0.85}
\renewcommand{\textfraction}{0.10}

\noindent This appendix is organized as follows.
\cref{sec:supp:def} details the operational verification probes behind the \emph{why}--\emph{when}--\emph{who} task taxonomy.
\cref{sec:supp:harness} describes the \harness (\texttt{TaskSpec}, acceptance checks, and a step-by-step dual-agent co-evolution case study).
\cref{sec:supp:bench} provides the benchmark comparison, task cards, prompts, and evaluation protocol of \bench.
\cref{sec:supp:model} gives implementation details of \method (token layout, joint causal mask, algorithms, training hyper-parameters, and real-time configuration).
\cref{sec:supp:baselines} describes the baseline training protocols, \cref{sec:supp:results} reports full simulation metrics and diagnostic analysis protocols, \cref{sec:supp:real} details the real-robot platform, assets, and experimental results, and \cref{sec:supp:more_cases} provides full-rollout qualitative comparisons and synchronized dual-robot imaginations across all seven \bench tasks.

\section{Task Definition: Verification Probes}
\label{sec:supp:def}

The \emph{why}--\emph{when}--\emph{who} labels in \cref{sec:bench:def} are operationally defined by deterministic verification probes executed in simulation by each task package's evaluator over the expert generator, rather than assigned by informal inspection.

\para{Necessity and complementarity (\emph{why}).}
Single-agent feasibility $\Feas(V;\Ecal)$ is probed by re-planning the task for a single robot with the task's expert planner under the unchanged goal predicate $V$.
A task satisfies multi-agent necessity (\cref{eq:necessity}) when the team expert succeeds while every single agent fails alone.
To identify the complementarity source $\kappa$ (\cref{eq:kappa}), the evaluator applies three single-agent relaxations in order of increasing physical substitution cost, $O\prec W\prec U$:
(i)~$\Rcal^{O}$ exposes the privileged state observable from the partner's cameras (\eg block colors on the far side of the table) while leaving reach and actuation unchanged;
(ii)~$\Rcal^{W}$ additionally permits re-placing the single agent's base across the union of both workspaces between motion segments;
(iii)~$\Rcal^{U}$ additionally grants the partner's effectors, payload, and simultaneous contact budget to one planner.
The first relaxation that restores single-agent feasibility determines $\kappa$: \emph{Information} ($\kappa{=}O$, \emph{Find common color}), \emph{Spatial} ($\kappa{=}W$, \emph{Joint bowl placing} and \emph{Hold and scan}), or \emph{Capability} ($\kappa{=}U$, the remaining four tasks).

\para{Temporal coupling (\emph{when}).}
To probe temporal overlap, the evaluator first \emph{serializes} the team expert so that agent $j$ finishes its critical phase and retreats to rest before agent $k$ begins.
Tasks that remain solvable under serialization are \emph{asynchronous} (\emph{Joint bowl placing}, \emph{Find common color}).
Otherwise, the evaluator performs a \emph{delay scan} by holding the first agent at its commanded pose while delaying the partner's critical phase by $\delta$:
if success persists across arbitrary waiting delays, the task is \emph{partially synchronous} (\emph{Hold and scan}, \emph{Open door, place pot});
if success breaks beyond a narrow tolerance $\delta^\star$ of a few control steps, an instant-by-instant physical constraint holds and the task is \emph{synchronous} (\emph{Lift tray, keep blocks}, \emph{Lift tray, pour food}, \emph{Pour and catch}).

\para{Embodiment (\emph{who}) and scope.}
A team is \emph{homogeneous} when both robots share identical URDFs, actuation spaces, and camera configurations, and \emph{heterogeneous} otherwise; for heterogeneous teams, a \emph{role-swap} probe verifies $\Feas(M_k;\Ecal_j)=0$, certifying that labor division is imposed by physical embodiment rather than language alone.
By design, \cref{eq:necessity} excludes tasks where a second robot merely accelerates a single-agent-feasible job (\eg parallel table clearing), where per-robot policies already suffice.
When a task contains multiple interaction segments, it is labeled by its strongest temporal coupling, and each \bench task is constructed around a single complementarity type for unambiguous evaluation.
Whereas classical multi-robot task-allocation taxonomies~\cite{gerkey2004mrta,korsah2013itax,nunes2017temporal} classify high-level assignment and precedence, our three axes characterize the physical necessity, timing constraints, and body heterogeneity faced directly by a learned low-level policy.

\section{\harness Details}
\label{sec:supp:harness}

\subsection{\texttt{TaskSpec}}
A \texttt{TaskSpec} is the single source of truth shared by the generation pipeline $G^{(n)}$ and evaluation pipeline $E^{(n)}$.
\cref{lst:taskspec} shows an abridged specification for \emph{Open door, place pot}.
Every predicate declared in \texttt{role\_milestones} and \texttt{collab\_milestones} is compiled into a deterministic, LLM-free detector over simulator state (contacts, poses, joint angles, and velocities); the exact same detectors gate demonstration acceptance and score learned policies.

\begin{figure}[t]
\begin{lstlisting}[caption={Abridged \texttt{TaskSpec} of \emph{Open door, place pot}. Thresholds marked \texttt{<..>} are refined by the harness during co-evolution.},label={lst:taskspec},captionpos=b]
task: open_door_place_pot
team:
  A: {robot: panda,  arms: 1, action_dim: 8,
      views: [head, wrist]}
  B: {robot: aloha,  arms: 2, action_dim: 16,
      views: [head, left_wrist, right_wrist]}
labels: {why: capability, when: partial,
         who: heterogeneous}
prompts:
  global: "Two robots put the pot into the
           cabinet together."
  A: "Open the cabinet door and hold it open."
  B: "Lift the pot with both hands and place
      it inside the cabinet."
goal V: in(pot, cabinet) and at_rest(pot)
role_milestones:
  A: [grasp(handle), door_angle >= <th_open>,
      release_after(pot_inside)]
  B: [bimanual_grasp(pot), lift(pot) >= <h>,
      in(pot, cabinet), retreat]
collab_milestones:
  - hold_during: {cond: door_angle >= <th_open>,
                  window: [B.enter, B.exit]}
  - no_contact: {bodies: [B.*, door]}
probes:
  necessity: [single_A, single_B,
              relax_O, relax_W, relax_U]
  temporal:  [serialize, delay: <deltas>]
  embodiment: [role_swap]
\end{lstlisting}
\end{figure}

\subsection{Acceptance Checks and Probes}
\para{Physical checks.}
A candidate demonstration is rejected if any of the following fails:
(i)~no robot--robot, robot--fixture, or self-collision outside whitelisted grasp contacts;
(ii)~inverse kinematics is continuous (no joint-branch jump between consecutive control steps) and strictly within joint limits;
(iii)~grasps form real finger--object contacts (both arms for bimanual holds) that persist throughout transport without pose overwriting or synthetic attachments;
(iv)~all objects remain physically supported and at rest in the terminal window.

\para{Logical checks and necessity probes.}
(v)~every role milestone in $M_k$ is reached in causal order within its critical interval $\phi_k$;
(vi)~every collaboration milestone in $\Mcollab$ holds across its entire shared window $\phi_{jk}$ (\eg the spring-loaded door stays above its angle threshold throughout insertion);
(vii)~the terminal goal predicate $V$ holds.
A trajectory is accepted iff (i)--(vii) all pass, and policy evaluation scores the exact same deterministic predicates (with IK continuity replaced by controller-level safety checks).
Once demonstrations pass acceptance, the evaluator runs the verification probes of \cref{sec:supp:def}; any mismatch against declared axis labels (\eg a weak door spring allowing single-agent completion) is fed back to the generation sub-agent as a scene-design failure.

\begin{figure*}[t]
  \centering
  \includegraphics[width=\textwidth]{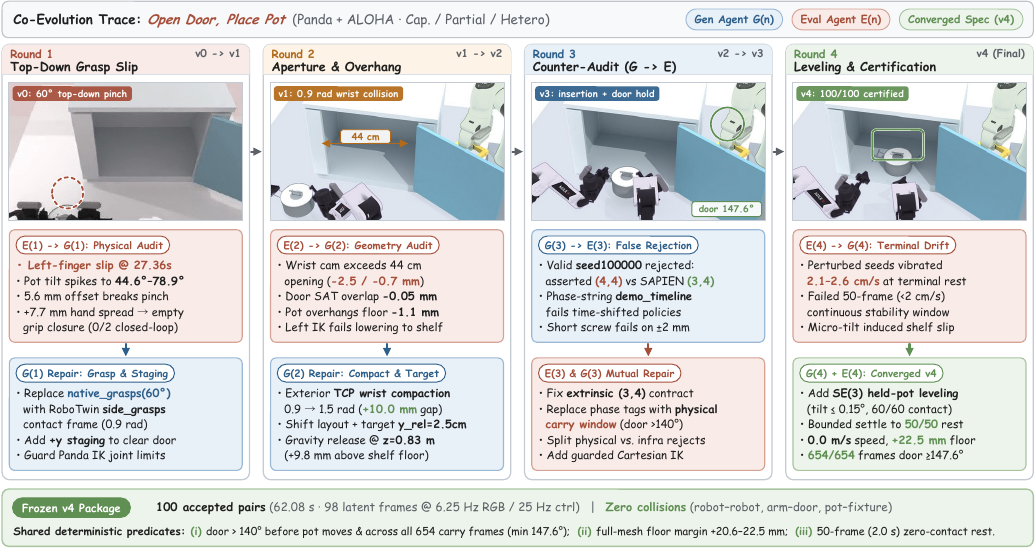}
  \vspace{-0.45em}
  \caption{\textbf{A detailed co-evolution example on \emph{Open door, place pot} (Panda\,+\,ALOHA).}
  Across four bidirectional rounds between the generation sub-agent $G^{(n)}$ (blue) and evaluation sub-agent $E^{(n)}$ (terracotta), physical audits repair top-down grasp slip, cabinet aperture interference, floor overhang, and terminal micro-oscillation, while a $G{\to}E$ counter-audit repairs false rejections from camera extrinsic shape mismatches and phase-string coupling, converging to the certified 100-pair v4 package.}
  \label{fig:supp:coevolution}
\end{figure*}

\subsection{A Detailed Example of Co-Evolution}
\cref{fig:supp:coevolution} illustrates a complete four-round co-evolution trace of the generation sub-agent $G^{(n)}$ and evaluation sub-agent $E^{(n)}$ on the heterogeneous task \emph{Open door, place pot}, where a single-arm Panda opens and holds a spring-return cabinet door (hinge stiffness $0.35$\,N$\cdot$m/rad, damping $0.6$) while a dual-arm ALOHA lifts a two-handled kitchen pot into the cabinet.

\para{Round~1 ($v_0{\to}v_1$): top-down grasp slip.}
Although the initial $60^\circ$ top-down pinch grasp (\texttt{native\_grasps}) produced nominal open-loop demonstrations, $E^{(1)}$'s closed-loop physical audit diagnosed severe millimetre fragility: under $5.6$\,mm hand offset and $+7.7$\,mm hand spread, the left fingers lost pot contact at $27.36$\,s, pot tilt spiked to $44.6^\circ$--$78.9^\circ$, and both fingers closed on empty air.
In response, $G^{(1)}$ replaced the top-down pinch with a RoboTwin contact-frame two-sided handle grasp (\texttt{side\_grasps} at $0.9$\,rad), introduced $+y$ staging to clear the held door, and added joint-limit filtering to the Panda IK solver.

\para{Round~2 ($v_1{\to}v_2$): aperture clearance and floor overhang.}
Auditing the new side-grasp trajectories, $E^{(2)}$ caught four geometric failures:
(i)~at $0.9$\,rad wrist orientation, ALOHA's wrist cameras exceeded the $44$\,cm cabinet opening by $-2.5$\,mm (left) and $-0.7$\,mm (right);
(ii)~lateral approach caused a $-0.05$\,mm SAT overlap with the held door at $147.6^\circ$;
(iii)~center-only placement at $y_{\text{rel}}{=}4.5$\,cm left the pot mesh overhanging the cabinet floor by $-1.1$\,mm;
(iv)~lowering directly to shelf height triggered left-arm IK failure.
$G^{(2)}$ added continuous exterior TCP wrist compaction ($0.9{\to}1.5$\,rad over 120 quintic steps, opening $+10.0$\,mm / $+11.5$\,mm aperture margins and $+6.5$\,mm door gap), shifted the layout and placement target to $y_{\text{rel}}{=}2.5$\,cm, and switched to low gravity release at $z{=}0.83$\,m ($+9.8$\,mm lowest-vertex clearance above the shelf floor).

\para{Round~3 ($v_2{\to}v_3$): bidirectional counter-audit ($G{\to}E$).}
Co-evolution is strictly bidirectional.
When $v_2$ rejected physically valid nominal trajectories (\eg \texttt{seed100000} with $+21.9$\,mm floor clearance and $0.0$\,m/s terminal speed), $G^{(3)}$ counter-audited $E^{(3)}$ and surfaced two evaluator bugs:
(i)~the schema validator asserted $(4,4)$ camera extrinsics whereas SAPIEN exports native $(3,4)$ matrices;
(ii)~legacy \texttt{demo\_timeline} checks matched hard-coded expert phase strings, falsely failing trajectories with shifted execution timing.
$E^{(3)}$ fixed the $(3,4)$ contract, separated physical rejections from infrastructure faults, and replaced phase strings with a phase-independent physical carry window (verifying \texttt{door\_angle} $>140^\circ$ and continuous Panda finger contact across every carry frame, $654/654$ frames, $\min 147.6^\circ$).
Simultaneously, $G^{(3)}$ added a bounded 50-step guarded Cartesian IK fallback for short screw-plan segments under $\pm 2$\,mm layout perturbations.

\para{Round~4 ($v_4$): held-pot leveling and convergence.}
Finally, $E^{(4)}$ flagged $2.1$--$2.6$\,cm/s terminal micro-oscillation on perturbed seeds that failed the 50-frame ($2.0$\,s) continuous rest check.
$G^{(4)}$ added exterior shared-$SE(3)$ held-pot leveling (reducing carry tilt to $\le 0.15^\circ$ under $60/60$ bilateral contact) and a bounded settling window until 50 consecutive zero-contact rest frames aligned to the 16-tick VAE grid.
The converged $v_4$ package froze 100 accepted demonstration pairs ($62.08$\,s, $98$ latent frames, zero collisions, full-mesh front-floor clearance $+20.6$--$22.5$\,mm, and $0.0$\,m/s terminal drift).

\FloatBarrier
\section{\bench Details}
\label{sec:supp:bench}

\begin{table}[t]
\centering\small\setlength{\tabcolsep}{3.1pt}
\caption{\textbf{Task-design coverage of collaboration benchmarks} (RoboTwin~2.0: bimanual; MangoBench: manipulation subset).}
\label{tab:bench_compare}
\vspace{2pt}
\begin{tabular}{@{}l cc ccc cc@{}}
\toprule
 & \multicolumn{2}{c}{When} & \multicolumn{3}{c}{Why} & \multicolumn{2}{c}{Who} \\
\cmidrule(lr){2-3}\cmidrule(lr){4-6}\cmidrule(l){7-8}
Benchmark & Asy. & Syn. & Info & Spa. & Cap. & Het. & Hom. \\
\midrule
RoCoBench~\cite{mandi2023roco}        & \cmark & \cmark & \cmark & \cmark & \cmark & \cmark & \xmark \\
TAPAS~\cite{zamora2024tapas}          & \cmark & \xmark & \xmark & \cmark & \xmark & \xmark & \cmark \\
Habitat-MAS~\cite{chen2024emos}       & \cmark & \xmark & \cmark & \cmark & \cmark & \cmark & \xmark \\
RoboFactory~\cite{qin2025robofactory} & \cmark & \cmark & \xmark & \cmark & \cmark & \xmark & \cmark \\
RoboTwin~2.0~\cite{chen2025robotwin2} & \cmark & \cmark & \xmark & \cmark & \cmark & \cmark & \cmark \\
MeCoBench~\cite{wang2026meco}         & \cmark & \cmark & \cmark & \cmark & \cmark & \cmark & \xmark \\
MangoBench~\cite{wang2026mangobench}  & \cmark & \cmark & \xmark & \xmark & \cmark & \xmark & \cmark \\
\midrule
\rowcolor{rowgray}\textbf{\bench (ours)} & \cmark & \cmark & \cmark & \cmark & \cmark & \cmark & \cmark \\
\bottomrule
\end{tabular}
\end{table}

\subsection{Simulation and Recording}
\cref{tab:bench_compare} compares \bench against existing collaborative benchmarks across the \emph{When}--\emph{Why}--\emph{Who} taxonomy.
\bench is built on RoboTwin~2.0~\cite{chen2025robotwin2} and the SAPIEN engine~\cite{xiang2020sapien}.
Physics runs at 250\,Hz and control at 25\,Hz.
Expert motions are planned in end-effector space and executed through a motion planner; learned policies are executed through the same controller interface.
Three embodiments are used: the dual-arm ALOHA-AgileX platform (two 6-DoF arms with parallel grippers), the 7-DoF Franka Panda, and the 6-DoF UR5.
Every robot records three RGB views.
A dual-arm robot records a head camera and two wrist cameras; a single-arm robot records a head (third-person) camera and its wrist camera, and its missing second wrist view is filled with a black image and masked out of the loss (\cref{sec:supp:model}).
Each demonstration pair stores, for both robots and on a common clock, the RGB streams, joint positions, end-effector poses, gripper states, native-dimension end-effector actions, the hierarchical prompts, and the full simulator state used by the evaluator.

\begin{table*}[t]
\centering\small\setlength{\tabcolsep}{4pt}
\caption{\textbf{\bench task cards.} A: ALOHA (dual-arm), P: Panda, U: UR5 (single-arm). ``Pairs'' counts accepted demonstration pairs; ``Len.'' is the nominal episode duration at 25\,Hz (aligned with \cref{fig:bench}).}
\label{tab:taskcards}
\vspace{-0.5em}
\begin{tabular}{@{}p{0.135\linewidth} c p{0.215\linewidth} p{0.205\linewidth} p{0.215\linewidth} c c@{}}
\toprule
Task {\scriptsize(When/Why/Who)} & Team & Role milestones $M_A$ & Role milestones $M_B$ & Collaboration milestones $\Mcollab$ & Pairs & Len.\,(s) \\
\midrule
Joint bowl placing {\scriptsize(Async/Spatial/Homo)} & A+A &
grasp bowl; place bowl in shared zone; retreat &
grasp bowl in shared zone; place bowl on goal pad; retreat &
bowl at rest in shared zone before B's grasp; no simultaneous grasp & 100 & 90 \\
\addlinespace[2pt]
Find common color {\scriptsize(Async/Info/Homo)} & A+A &
touch only the shared-color block; place it on own pad &
touch only the shared-color block; place it on own pad &
both placed blocks have the same color; no other block displaced & 100 & 13 \\
\addlinespace[2pt]
Hold and scan {\scriptsize(Partial/Spatial/Homo)} & A+A &
grasp box; lift and present it in shared zone; return box &
grasp scanner; align scanner with box; return scanner &
scanner within distance and angle tolerance of the held box for $\ge 0.6$\,s & 100 & 56 \\
\addlinespace[2pt]
Open door, place pot {\scriptsize(Partial/Cap./Hetero)} & P+A &
grasp handle; open door beyond threshold; release after pot is inside &
bimanual grasp of pot handles; lift pot; pot inside cabinet; retreat &
door held open during the whole insertion window; no arm--door contact & 100 & 62 \\
\addlinespace[2pt]
Lift tray, keep blocks {\scriptsize(Sync/Cap./Homo)} & A+A &
two-hand grasp of own tray side; hold through transport; release after set-down &
(same, mirrored) &
four-gripper contact during transport; tray tilt within bound; load stays on tray; load at rest after set-down & 100 & 40 \\
\addlinespace[2pt]
Lift tray, pour food {\scriptsize(Sync/Cap./Homo)} & A+A &
two-hand grasp of own tray side; hold through transport and tilt &
(same, mirrored) &
four-gripper contact; synchronized tilt of about $8^\circ$ at the bag; all three items in the bag and at rest & 100 & 39 \\
\addlinespace[2pt]
Pour and catch {\scriptsize(Sync/Cap./Hetero)} & P+U &
grasp hopper; tilt hopper to pour; return hopper &
grasp cup; raise cup under outlet; set cup on target &
cup under the moving outlet throughout pouring; all balls caught & 100 & 38 \\
\midrule
Total & & & & & 700 & \\
\bottomrule
\end{tabular}
\end{table*}

\subsection{Task Cards}
\cref{tab:taskcards} lists the milestones and nominal episode durations of all seven tasks, matching \cref{fig:bench}.
Key task-specific details follow.

\para{Find common color.}
The bases of the two robots are $2.7$\,m apart at opposite ends of a long table, so neither robot's cameras see the partner's blocks.
Each robot faces three blocks drawn so that the two sets share exactly one color.
The local role instructions never name the color (both robots receive symmetric text up to the robot ID), so the shared color can only be inferred from the partner's observation.

\para{Lift tray, keep blocks.}
Demonstrations are generated with a single free-rolling ball on the tray, forcing the expert to level the tray actively before lifting and set-down; acceptance requires four-gripper contact during transport and rest ($\le 1$\,mm/s) during the final 2\,s.
At evaluation time, the ball is replaced by stacked $4$\,cm, $40$\,g wooden cubes on the tray; all blocks must remain upright on the tray after set-down.

\para{Lift tray, pour food.}
The tray carries an apple, a carrot, and a cucumber.
After lifting, both robots transport the tray to the table edge and tilt it in unison by about $8^\circ$ over an open bag, maintaining contact until all three items settle inside.

\para{Open door, place pot.}
The cabinet door is spring-loaded and snaps shut within a fraction of a second when released, while the pot requires a bimanual two-sided handle grasp; no single robot can simultaneously hold the door and insert the pot.

\para{Pour and catch.}
The hopper outlet traces an arc as the Panda tilts it, so the UR5 must continuously track the moving outlet with the cup throughout pouring.

\subsection{Instructions}
Each task provides one global team instruction $g$ and one role instruction $\ell^k$ per robot.
The stored prompts are written in Chinese; English translations of two representative tasks are as follows.
For \emph{Lift tray, pour food}: $g$ = ``The two robots put everything into the garbage bag together.''; $\ell^A$ = ``You are robot A. Grasp the side of the tray near you with both arms; the tray holds an apple, a carrot and a cucumber. Together with robot B, lift the tray, move it to the table edge, tilt it to pour all three items into the open bag, and keep holding the tray until the food has settled.'' ($\ell^B$ is symmetric.)
For \emph{Open door, place pot}: $g$ = ``Two robots put the pot into the cabinet together: the single arm opens the door and the dual arm places the pot.''; $\ell^{A}$ = ``Open the cabinet door and hold it open.''; $\ell^{B}$ = ``Lift the pot with both hands and place it inside the cabinet.''

\subsection{Training Data and Evaluation Protocol}
Training uses the 100 accepted demonstration pairs per task listed in \cref{tab:taskcards}.
For evaluation, we perform \textbf{100 independent trajectory trials per task} directly in the simulator: each trial instantiates a fresh scene configuration (randomized object poses, colors, and identities distinct from the training demonstration trajectories) and rolls out the learned policy in closed loop under the task's deterministic evaluator.
Before benchmarking policies, we verify that the scripted expert solves the evaluation layouts.
Because joint success is binary, we report 95\% Wilson score confidence intervals alongside point estimates (\cref{sec:supp:results}).

\FloatBarrier
\section{\method Implementation Details}
\label{sec:supp:model}

\subsection{Token Layout and Masks}
\para{Video tokens.}
Each robot's three RGB views are resized and tiled onto one canvas before VAE encoding: the head view occupies the top $16{\times}20$ latent cells and the two wrist views occupy the two $8{\times}10$ cells below it, producing a latent $z^k_f\in\mathbb{R}^{48\times 24\times 20}$ per latent frame ($48$ channels) that is patchified with a $1{\times}2{\times}2$ kernel into $120$ tokens.
Any view absent for a single-arm embodiment is encoded from a black image and excluded from the video loss by a spatial view mask.
Video is sampled at stride 4 from the 25\,Hz control stream (6.25\,Hz) and the causal video VAE compresses time by $4\times$, so one latent frame spans 16 control steps.

\para{Action tokens.}
Actions are end-effector targets: per arm, a 3-D position, unit quaternion, and gripper command.
The backbone's shared $30$-D action slot layout places two 7-D arm blocks at channels $0$--$6$ and $7$--$13$ and two gripper channels at $28$ and $29$.
Dual-arm robots (ALOHA) use channels $\{0,\dots,13,28,29\}$ (16 active); single-arm robots (Panda, UR5) use $\{0,\dots,6,28\}$ (8 active).
Inactive channels are zeroed in clean actions, injected noise, noisy/conditioning inputs, flow targets, and losses, and are re-zeroed after every denoising step at inference; without re-zeroing, noise in unused slots leaks through attention and causes jitter.
Poses are expressed relative to the first-frame end-effector pose, normalized to $[-1,1]$ via per-task $1\%$/$99\%$ quantiles, and clipped.
Each control step forms one action token (16 action tokens per latent frame).

\para{Sequence.}
For robot $k$ and latent frame $f$, the sequence comprises four token groups: noisy video, clean video, noisy action, and clean action.
Over $F$ latent frames, each robot contributes $L=2F(120+16)$ tokens during training and the joint sequence has length $N{\cdot}L$.
All robots share identical 3-D rotary coordinates (frame, height, width) and are distinguished solely by the fixed identity and role encodings of \cref{sec:method:arch}.

\begin{figure*}[t]
  \centering
  \includegraphics[width=\textwidth]{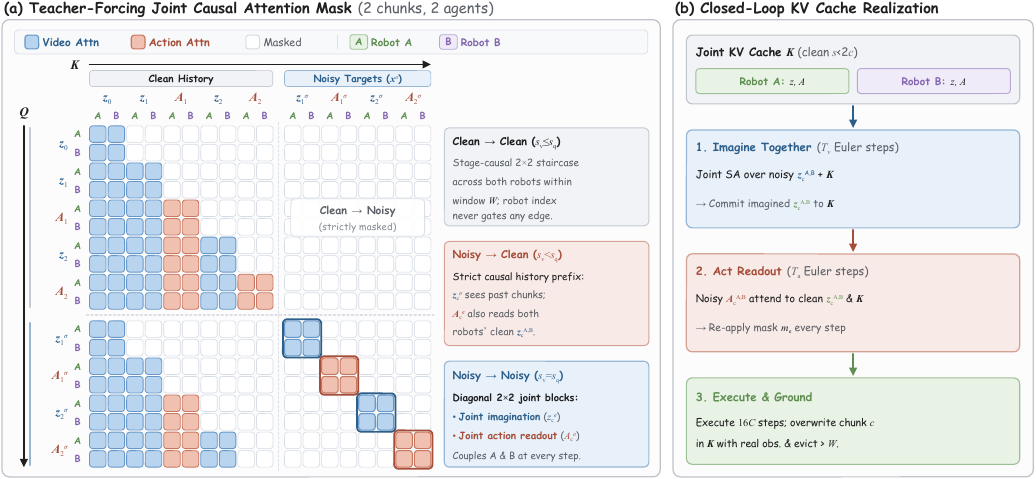}
  \vspace{-0.45em}
  \caption{\textbf{Teacher-forcing joint causal attention mask and closed-loop KV cache realization.}
  \textbf{(a)}~Stage-interleaved $Q\times K$ attention mask (\cref{eq:mask}) for two chunks and two robots ($A, B$): clean history follows a $2{\times}2$ stage-causal lower-triangular staircase ($s_v{\le}s_q$) with clean-to-noisy strictly masked; noisy targets attend strictly to earlier clean stages ($s_v{<}s_q$, so noisy actions $A_c^\sigma$ read both robots' clean video $z_c^{A,B}$) and couple across robots via diagonal $2{\times}2$ blocks ($s_v{=}s_q$) for joint imagination and joint action readout.
  \textbf{(b)}~At inference, the mask is realized by two sequential denoising stages over a sliding joint KV cache $\mathcal{K}$: noisy future video latents are denoised jointly first and committed as clean video tokens, after which noisy actions attend to all robots' imagined video latents before execution and real-observation grounding.}
  \label{fig:supp:mask}
\end{figure*}

\subsection{Joint Causal Attention Mask}
\cref{fig:supp:mask}a visualizes the teacher-forcing joint causal attention mask (\cref{eq:mask}) over two chunks and two robots ($A, B$), interleaving agents within each stage to reveal its four-quadrant block structure.
Because the robot index never gates an attention edge, every causal stage forms a symmetric $2{\times}2$ cross-agent block: clean history builds a lower-triangular staircase, noisy targets see only strictly earlier clean stages (allowing noisy actions $A_c^\sigma$ to read both robots' clean video $z_c^{A,B}$), and same-stage noisy tokens interact along diagonal $2{\times}2$ blocks for joint imagination and joint action readout.
During training, chunk size and history window are randomized over $C\in\{1,\dots,4\}$ and $W\in\{4,\dots,64\}$, and with probability $0.5$ the clean video history receives light noise to improve robustness against deployment drift.
At inference (\cref{fig:supp:mask}b), the clean $z_c^{A,B}$ keys are populated by the model's own jointly denoised video latents before action denoising begins (step 2 of \cref{alg:infer}), realizing the exact block mask via the sliding joint KV cache $\mathcal{K}$.

\subsection{Algorithms}
\cref{alg:block} details one weight-shared \method transformer block and \cref{alg:infer} summarizes the four-step closed-loop inference procedure.
The first latent frame of the initial chunk is pinned to the encoded initial observation, so the first chunk outputs 16 control steps and each subsequent chunk outputs $16C=32$ steps.
Classifier-free guidance is applied solely during the video denoising loop.

\begin{algorithm}[t]
\caption{One \method block $b$ (weights shared across robots).}
\label{alg:block}
\small
\begin{algorithmic}[1]
\Require hidden states $h=[h_1;\dots;h_N]$, timestep embedding $\tau$, goal tokens $G$, role tokens $\{L_k\}$, joint causal mask $\mathsf{M}$ (\cref{eq:mask})
\State $\tilde h \gets h + \SA_b\big(\mathrm{AdaLN}_b(h,\tau);\ \mathsf{M}\big)$ \Comment{joint over all robots}
\State $h' \gets \tilde h + \gamma_b\,\CA^{g}_b\big(\LN^{g}_b(\tilde h),\,G\big)$ \Comment{global goal, gated}
\For{$k=1,\dots,N$}
  \State $h''_k \gets h'_k + \CA^{\ell}_b\big(\LN_b(h'_k),\,L_k\big)$ \Comment{own role only}
\EndFor
\State \Return $h'' + \FFN_b\big(\mathrm{AdaLN}'_b(h'',\tau)\big)$
\end{algorithmic}
\end{algorithm}

\begin{algorithm}[t]
\caption{Closed-loop joint inference of \method.}
\label{alg:infer}
\small
\begin{algorithmic}[1]
\Require initial observations $\{o^k_0\}$, instructions $g$, $\{\ell^k\}$, chunk size $C$, window $W$, steps $(T_v,T_a)$, guidance scale $w$
\State encode text once; encode $\{o^k_0\}$ with per-robot streaming VAE states; write them to the joint KV cache $\mathcal{K}$ at $\sigma{=}0$
\While{task not finished}
  \State $\{\hat z^k_c\}_{k=1}^{N} \sim \mathcal{N}(0,I)$ \Comment{future video latents of all robots}
  \For{$i=1,\dots,T_v$} \Comment{(1) imagine jointly}
     \State $\{\hat z^k_c\}\gets \textsc{EulerStep}\big(\theta,\{\hat z^k_c\},\mathcal{K},g,\{\ell^k\};w\big)$
  \EndFor
  \State write $\{\hat z^k_c\}$ to $\mathcal{K}$ as clean video tokens of chunk $c$
  \State $\{\hat A^k_c\}_{k=1}^{N} \sim \mathcal{N}(0,I)$;\ \ $\hat A^k_c\gets m_{e_k}\odot\hat A^k_c$
  \For{$i=1,\dots,T_a$} \Comment{(2) act}
     \State $\{\hat A^k_c\}\gets \textsc{EulerStep}\big(\theta,\{\hat A^k_c\},\mathcal{K},g,\{\ell^k\};1\big)$
     \State $\hat A^k_c\gets m_{e_k}\odot\hat A^k_c$ for all $k$ \Comment{re-apply masks}
  \EndFor
  \State de-normalize $\{\hat A^k_c\}$, map slots to native commands, execute $16C$ control steps on all robots \Comment{(3) execute}
  \State encode new observations $\{o^k\}$; overwrite chunk $c$ in $\mathcal{K}$ with real latents and executed actions at $\sigma{=}0$ \Comment{(4) ground}
  \State evict tokens older than $W$ latent frames from $\mathcal{K}$
\EndWhile
\end{algorithmic}
\end{algorithm}

\subsection{Training Details}
\cref{tab:hparams} summarizes the training and inference hyper-parameters.
All tasks are trained on \textbf{8$\times$ NVIDIA H20 GPUs} using FSDP in \texttt{bfloat16} precision with activation checkpointing and a global batch size of 8 full episodes (one episode per GPU).
Each task undergoes \textbf{3,000 updates in Stage~1} ($N{=}1$, whole single-robot trajectories with a 1:1 embodiment ratio on heterogeneous teams) followed by \textbf{3,000 updates in Stage~2} ($N{=}2$, whole synchronized multi-robot episodes initialized from the task's Stage~1 checkpoint).
Losses are normalized over active action slots and valid camera views before averaging across samples so single-arm robots are not diluted by masked channels, and per-robot video/action losses are logged separately in Stage~2.
No embodiment-specific parameters are introduced in either stage.

\begin{table}[t]
\centering\small\setlength{\tabcolsep}{4pt}
\caption{\textbf{Training and inference hyper-parameters.}}
\label{tab:hparams}
\vspace{-0.4em}
\resizebox{\linewidth}{!}{%
\begin{tabular}{@{}l l@{}}
\toprule
\multicolumn{2}{@{}l}{\emph{Model}}\\
Backbone & \lingbot~\cite{li2026lingbotva}, 30 blocks, $\approx$5\,B params \\
Text encoder / VAE & frozen T5 / frozen causal video VAE \\
New parameters & global cross-attention + norm + gate per block \\
Identity / role gain $\beta$ & $0.01$ (fixed Fourier ID, no learned table) \\
Gate init.\ $\gamma_b$ & $10^{-3}$ \\
\midrule
\multicolumn{2}{@{}l}{\emph{Optimization (both stages)}}\\
Hardware & 8$\times$ NVIDIA H20 GPUs \\
Optimizer & AdamW, $\beta{=}(0.9,0.95)$, weight decay $0.1$ \\
Learning rate & $10^{-5}$ constant; $10^{-4}$ (no decay) for $\gamma_b$ \\
Batch size & 8 episodes (1 per GPU across 8 GPUs) \\
Gradient clipping & $2.0$ \\
Precision / memory & bfloat16, FSDP, activation checkpointing \\
Noise schedule & 1000 levels; shift 5 (video), 1 (action) \\
Text dropout (CFG) & $0.1$, all instructions dropped jointly \\
Chunk / window (train) & $C\in\{1..4\}$, $W\in\{4..64\}$, randomized \\
Stage~1 / Stage~2 updates & 3,000 / 3,000 steps per task \\
\midrule
\multicolumn{2}{@{}l}{\emph{Inference}}\\
Chunk / window & $C{=}2$ (32 actions), $W{=}64$ \\
Steps (benchmark) & 25 video (guidance 5), 50 action (no guidance) \\
Steps (real-time) & 4 video, 8 action \\
\bottomrule
\end{tabular}}
\end{table}

\subsection{Real-Time Configuration}
Consistent with \cref{sec:method:infer,sec:supp:real}, our real-time deployment modifies only the sampler and attention kernels without altering model weights:
(i)~4 video and 8 action Euler steps paired with asynchronous parallel decoding;
(ii)~fused QKV linear projections and a compiled 3-D RoPE kernel in every self-attention block;
(iii)~episode-level caching of global and local text keys and values;
(iv)~sliding joint KV cache with CUDA Graphs so each denoising step processes only the current chunk.
On a single GPU, this reduces median chunk inference latency from $3.92$\,s (25/50 steps) to $0.72$\,s ($5.4\times$ speedup), enabling pause-free closed-loop real-robot execution (\cref{sec:supp:real}).

\FloatBarrier
\section{Baselines}
\label{sec:supp:baselines}

\begin{table}[t]
\centering\small\setlength{\tabcolsep}{3pt}
\caption{\textbf{Baseline training configurations.} Views: cameras per policy instance; Fut.: predicts future observations; X-attn: cross-robot coupling; Updates: per-task fine-tuning steps (S1+S2 where two-stage).}
\label{tab:baseline_cfg}
\vspace{-0.4em}
\resizebox{\linewidth}{!}{%
\begin{tabular}{@{}l c c c c c@{}}
\toprule
Method & Instances & Views & Fut. & X-attn & Updates \\
\midrule
\pizf (per-robot)    & $N$ & 3 & \xmark & \xmark & 3k \\
\lingbot (per-robot) & $N$ & 3 & \cmark & \xmark & 3k (S1) \\
\pizf (joint)        & 1 & 6 & \xmark & flat & 3k + 3k \\
\lingbot (joint)     & 1 & 6 & \cmark & flat & 3k + 3k \\
MA-VLA               & 1 & 6 & \xmark & prompt & 6k \\
VLA-Fabric           & per-arm & 2/arm & \xmark & peer & 6k \\
\rowcolor{rowgray}\textbf{\method (ours)} & 1 & $3N$ & \cmark & agent axis & 3k + 3k \\
\bottomrule
\end{tabular}}
\end{table}

\begin{table*}[t]
\centering\small\setlength{\tabcolsep}{3.2pt}
\caption{\textbf{Full metrics on \bench}: Role Scores of both robots $S_A$/$S_B$, Collaboration Score $\Scollab$ (\%), and joint Success Rate $\SR$ (\%) with 95\% Wilson confidence intervals across all seven methods over 100 independent trajectory trials per task.}
\label{tab:full}
\vspace{-0.5em}
\resizebox{\textwidth}{!}{%
\begin{tabular}{@{}l ccc ccc ccc ccc@{}}
\toprule
 & \multicolumn{3}{c}{\pizf (per-robot)} & \multicolumn{3}{c}{\lingbot (per-robot)} & \multicolumn{3}{c}{\pizf (joint)} & \multicolumn{3}{c}{\lingbot (joint)} \\
\cmidrule(lr){2-4}\cmidrule(lr){5-7}\cmidrule(lr){8-10}\cmidrule(l){11-13}
Task & $S_A$/$S_B$ & $\Scollab$ & $\SR$ [CI] & $S_A$/$S_B$ & $\Scollab$ & $\SR$ [CI] & $S_A$/$S_B$ & $\Scollab$ & $\SR$ [CI] & $S_A$/$S_B$ & $\Scollab$ & $\SR$ [CI] \\
\midrule
Joint bowl placing     & 63.5/2.5 & 4.0 & 0.0\,[0.0,3.7]   & 65.0/6.0   & 6.5  & 5.0\,[2.2,11.2]  & 49.5/0.0 & 0.0 & 0.0\,[0.0,3.7] & 62.0/14.0   & 16.0 & 5.0\,[2.2,11.2] \\
Find common color      & 57.0/56.5& 8.5 & 6.0\,[2.8,12.5]  & 98.5/99.0  & 12.5 & 11.0\,[6.3,18.6] & 0.5/0.5  & 0.0 & 0.0\,[0.0,3.7] & 98.5/97.5   & 30.5 & 28.0\,[20.1,37.5] \\
Hold and scan          & 40.5/32.0& 30.5& 26.0\,[18.4,35.4]& 91.0/95.0  & 90.5 & 82.0\,[73.3,88.3]& 0.0/0.0  & 0.0 & 0.0\,[0.0,3.7] & 48.5/0.0    & 0.0  & 0.0\,[0.0,3.7] \\
Open door, place pot   & 56.5/29.0& 29.0& 21.0\,[14.2,30.0]& 100.0/90.5 & 90.5 & 81.0\,[72.2,87.5]& 0.0/0.0  & 0.0 & 0.0\,[0.0,3.7] & 50.5/0.5    & 0.5  & 0.0\,[0.0,3.7] \\
Lift tray, keep blocks & 7.5/7.5  & 2.3 & 0.0\,[0.0,3.7]   & 66.5/66.5  & 59.0 & 47.0\,[37.5,56.7]& 0.0/0.0  & 0.0 & 0.0\,[0.0,3.7] & 0.0/0.0     & 0.0  & 0.0\,[0.0,3.7] \\
Lift tray, pour food   & 21.5/21.5& 11.5& 0.0\,[0.0,3.7]   & 85.0/85.5  & 67.0 & 41.0\,[31.9,50.8]& 0.0/0.0  & 0.0 & 0.0\,[0.0,3.7] & 100.0/100.0 & 81.5 & 63.0\,[53.2,71.8] \\
Pour and catch         & 25.5/19.0& 20.0& 17.0\,[10.9,25.5]& 100.0/97.5 & 98.0 & 97.0\,[91.5,99.0]& 10.0/1.0 & 0.5 & 0.0\,[0.0,3.7] & 1.5/0.0     & 0.0  & 0.0\,[0.0,3.7] \\
\midrule
 & \multicolumn{3}{c}{MA-VLA} & \multicolumn{3}{c}{VLA-Fabric} & \multicolumn{3}{c}{\textbf{\method (ours)}} & \multicolumn{3}{c}{} \\
\cmidrule(lr){2-4}\cmidrule(lr){5-7}\cmidrule(lr){8-10}
Task & $S_A$/$S_B$ & $\Scollab$ & $\SR$ [CI] & $S_A$/$S_B$ & $\Scollab$ & $\SR$ [CI] & $S_A$/$S_B$ & $\Scollab$ & $\SR$ [CI] & & & \\
\midrule
Joint bowl placing     & 60.5/2.0 & 3.0 & 0.0\,[0.0,3.7]   & 68.0/13.5 & 13.5 & 9.0\,[4.8,16.2]  & 93.0/73.5   & 77.0 & 66.0\,[56.3,74.5] & & & \\
Find common color      & 84.5/83.5& 37.5& 35.0\,[26.4,44.7]& 87.5/86.5 & 35.5 & 33.0\,[24.6,42.7]& 100.0/100.0 & 66.5 & 63.0\,[53.2,71.8] & & & \\
Hold and scan          & 35.0/5.0 & 2.0 & 0.0\,[0.0,3.7]   & 69.0/52.0 & 47.5 & 42.0\,[32.8,51.8]& 97.5/98.5   & 97.0 & 95.0\,[88.8,97.8] & & & \\
Open door, place pot   & 73.0/40.0& 40.0& 31.0\,[22.8,40.6]& 43.0/4.5  & 4.5  & 0.0\,[0.0,3.7]   & 100.0/94.5  & 94.5 & 88.0\,[80.2,93.0] & & & \\
Lift tray, keep blocks & 7.5/7.5  & 2.3 & 0.0\,[0.0,3.7]   & 12.0/12.0 & 3.7  & 0.0\,[0.0,3.7]   & 95.5/95.5   & 93.7 & 93.0\,[86.3,96.6] & & & \\
Lift tray, pour food   & 85.0/85.0& 61.0& 38.0\,[29.1,47.8]& 86.5/87.0 & 61.0 & 36.0\,[27.3,45.8]& 100.0/100.0 & 95.5 & 91.0\,[83.8,95.2] & & & \\
Pour and catch         & 13.0/1.5 & 2.0 & 0.0\,[0.0,3.7]   & 14.0/1.5  & 2.0  & 0.0\,[0.0,3.7]   & 87.0/84.5   & 86.0 & 83.0\,[74.5,89.1] & & & \\
\bottomrule
\end{tabular}}
\end{table*}

All baselines are trained per task on the identical 100 demonstration pairs per task and evaluated over the same 100 independent simulator trajectory trials with the same end-effector action interface and deterministic evaluator (\cref{tab:baseline_cfg}).

\para{Per-robot deployment.}
Each robot runs an independent policy instance receiving only its own three views, history, and role instruction $\ell^k$, predicting only its own actions concurrently in closed loop (no partner camera access and no expert-replay partner).
\emph{\lingbot (per-robot)} directly uses our Stage~1 checkpoint (trained for 3,000 steps on per-robot trajectories with embodiment masks; one shared weight set controls both robots).
\emph{\pizf (per-robot)} fine-tunes \pizf~\cite{black2025pi05} on the exact same per-robot trajectories for the same 3,000 steps using the corresponding active slices of its action vector.
For \emph{Find common color}, the local role prompt never reveals the target color for any method.

\para{Joint deployment with all views.}
One centralized policy takes all six views and the concatenated global and role instructions as a single flat stream, predicting both robots' concatenated actions.
Both \emph{\pizf (joint)} and \emph{\lingbot (joint)} continue training for 3,000 steps from their respective Stage~1 checkpoints, directly aligned with \method's two-stage schedule; in particular, \emph{\lingbot (joint)} initializes from the exact same Stage~1 checkpoint as \method and trains on the synchronized paired demonstrations with the standard \lingbot video--action objective (tiling six views into the video canvas and packing both robots' active channels into the action vector).

\para{Existing multi-robot methods.}
\emph{MA-VLA}~\cite{zhang2026mavla} follows its released decomposed-prompt architecture using our global and role instructions as task and per-arm prompts.
\emph{VLA-Fabric}~\cite{zhang2026vlafabric} instantiates one arm-local VLA policy per physical arm on its robot's head view, wrist view, and proprioception, coupled through learned peer-representation interaction across arms.
Because both methods are single-stage multi-robot policies without a separate Stage~1 pre-adaptation phase, each is trained for \textbf{6,000 steps} per task on the synchronized multi-robot demonstrations to match the combined two-stage step budget ($3\text{k}+3\text{k}$) of the two-stage methods.

\section{Additional Simulation Results}
\label{sec:supp:results}

\subsection{Full Metrics and Confidence Intervals}
\cref{tab:full} reports per-robot Role Scores $S_A$/$S_B$, Collaboration Score $\Scollab$, and joint Success Rate $\SR$ with 95\% Wilson confidence intervals for all seven methods.
Per-robot policies maintain relatively high Role Scores on asynchronous and partially synchronous tasks while dropping sharply in Collaboration Score; on synchronous tasks even Role Scores degrade once partner trajectories diverge during coupled transport.

\subsection{Breakdown by Axis}

\begin{table}[b]
\centering\small\setlength{\tabcolsep}{3.2pt}
\caption{\textbf{Joint success (\%) grouped by axis value} (mean over the tasks carrying each label).}
\label{tab:axes}
\vspace{-0.4em}
\resizebox{\linewidth}{!}{%
\begin{tabular}{@{}l ccc ccc cc@{}}
\toprule
 & \multicolumn{3}{c}{When} & \multicolumn{3}{c}{Why} & \multicolumn{2}{c}{Who} \\
\cmidrule(lr){2-4}\cmidrule(lr){5-7}\cmidrule(l){8-9}
Method & Asy. & Par. & Syn. & Info & Spa. & Cap. & Hom. & Het. \\
\midrule
\pizf (per-robot)    &  3.0 & 23.5 &  5.7 &  6.0 & 13.0 &  9.5 &  6.4 & 19.0 \\
\lingbot (per-robot) &  8.0 & 81.5 & 61.7 & 11.0 & 43.5 & 66.5 & 37.2 & \textbf{89.0} \\
\pizf (joint)        &  0.0 &  0.0 &  0.0 &  0.0 &  0.0 &  0.0 &  0.0 &  0.0 \\
\lingbot (joint)     & 16.5 &  0.0 & 21.0 & 28.0 &  2.5 & 15.8 & 19.2 &  0.0 \\
MA-VLA               & 17.5 & 15.5 & 12.7 & 35.0 &  0.0 & 17.3 & 14.6 & 15.5 \\
VLA-Fabric           & 21.0 & 21.0 & 12.0 & 33.0 & 25.5 &  9.0 & 24.0 &  0.0 \\
\rowcolor{rowgray}\textbf{\method} & \textbf{64.5} & \textbf{91.5} & \textbf{89.0} & \textbf{63.0} & \textbf{80.5} & \textbf{88.8} & \textbf{81.6} & \underline{85.5} \\
\bottomrule
\end{tabular}}
\end{table}

\cref{tab:axes} regroups \cref{tab:main} by axis value.
Over centralized and multi-robot VLA baselines, \method's margin expands from asynchronous ($+43.5$) to synchronous and partially synchronous tasks ($+68.0$ and $+70.5$), from information ($+28.0$) to spatial ($+55.0$) and capability complementarity ($+71.5$), and from homogeneous ($+57.6$) to heterogeneous teams ($+70.0$).

\subsection{Diagnostic Analysis Details}
\label{sec:supp:consistency}
Below we specify the exact experimental design and metric computation for the three diagnostic probes in \cref{sec:exp:analysis} (\cref{fig:analysis}).

\begin{figure*}[t]
  \centering
  \includegraphics[width=\textwidth]{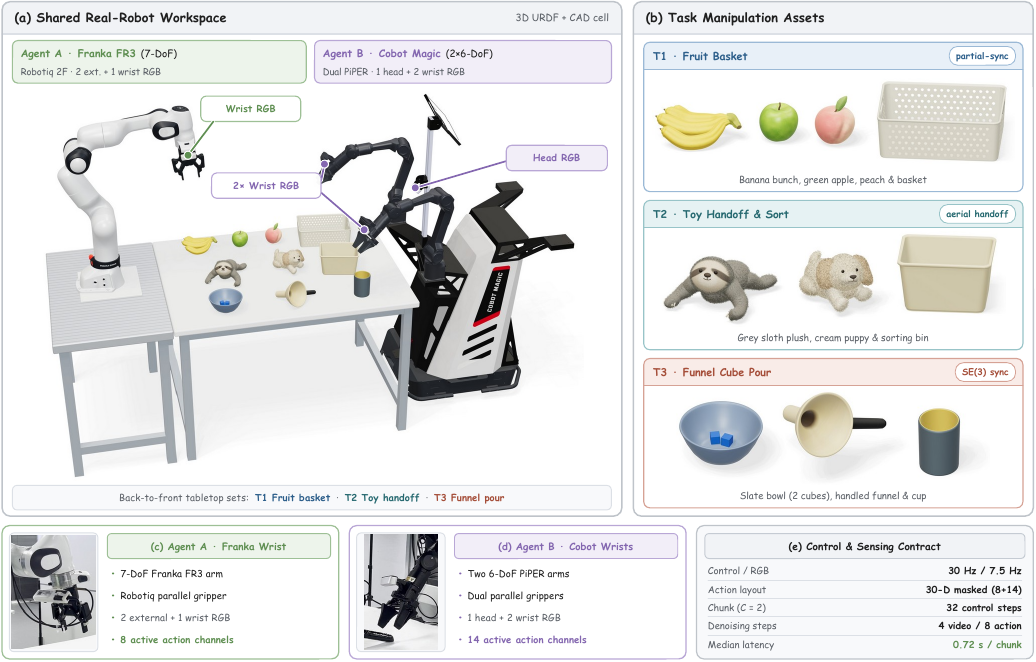}
  \vspace{-0.45em}
  \caption{\textbf{Unified real-robot hardware cell, task manipulation assets, and runtime contract.}
  \textbf{(a)}~3D URDF\,+\,CAD reconstruction of the shared workspace pairing Agent~A (7-DoF Franka FR3 with Robotiq parallel gripper and wrist RGB) opposite Agent~B (dual-arm AgileX Cobot Magic with two 6-DoF PiPER arms, dual wrist RGB, and mast-mounted head RGB), with all three task sets arranged from back to front.
  \textbf{(b)}~Close-up 3D manipulation assets for \emph{Fruit basket} (T1), \emph{Toy handoff \& sort} (T2), and \emph{Funnel cube pour} (T3).
  \textbf{(c--e)}~Physical wrist/end-effector camera modules and the shared closed-loop control and sensing contract.}
  \label{fig:supp:real_assets}
\end{figure*}

\para{(1) Partner-cue counterfactual protocol.}
To test whether a policy genuinely conditions on its partner's visual observations, we evaluate 25 unseen scenes of \emph{Find common color} as 50 paired directional trials (treating Robot~A and Robot~B as the focal robot in turn).
For each scene, we construct a counterfactual twin by recoloring \emph{only} the partner's three blocks so that the unique shared color shifts from the original focal slot $s_{\mathrm{orig}}$ to a distinct slot $s_{\mathrm{new}}\neq s_{\mathrm{orig}}$, while keeping the focal robot's blocks, poses, and $t{=}0$ three-camera RGB inputs strictly bit-identical ($\max|\Delta I_0|{=}0$).
We measure the \emph{target flip rate}—the fraction of originally solved trials in which the focal robot changes its placed block to $s_{\mathrm{new}}$—and overall counterfactual target accuracy across all 50 trials.

\para{(2) 3-D cross-robot imagination disagreement.}
To test whether both robots' imagined futures agree in 3-D physical space, we roll out six consecutive open-loop chunks ($12$ latent frames $= 48$ RGB frames $= 7.68$\,s) from identical history prefixes and identical Gaussian noise draws $(\epsilon_A,\epsilon_B)$ across 20 unseen scenes (\texttt{episodes 80..99}) of \emph{Lift tray, keep blocks} and \emph{Lift tray, pour food}, and decode each robot's imagined latents offline with the frozen causal Wan2.2-VAE (no decoding occurs during control).
To estimate world-frame 3-D tray pose without cross-view leakage, we train a single-view CoordConv-ResNet18 6-DoF pose probe $(I_k,\mathrm{cam\_id}_k)\mapsto(\hat p^{\,k},\hat R^{\,k})$ solely on ground-truth VAE-decoded single-view head frames from training episodes \texttt{0..79} (never exposing paired views or generated frames during probe training; validation MAE is $1.46$\,mm / $0.40^\circ$ on \emph{keep blocks} with a $1.40$\,mm paired-camera noise floor, and $0.89$\,mm / $0.24^\circ$ on \emph{pour food} with a $0.41$\,mm floor).
The \emph{cross-robot imagination disagreement} at horizon $h$ is computed as the Euclidean distance between the two robots' predicted world-frame tray centers,
\begin{equation}
  D(h)=\big\|\,\hat p^{\,A}_{t+h}-\hat p^{\,B}_{t+h}\big\|_2 ,
\end{equation}
averaged over the 20 unseen episodes.
We further test active reconciliation under a $0.64$\,s ($1$-latent-frame) partner history phase offset injected before open-loop rollout.

\para{(3) Partner delay protocol.}
To evaluate closed-loop re-synchronization under partner lag, we run 20 unseen scenes of \emph{Lift tray, pour food} while holding Robot~B's initial joint and gripper command via zero-order hold for $\delta\in\{0.00,0.32,0.64,0.96,1.28,1.60,1.92,2.56\}$\,s ($\mathrm{round}(25\delta)$ control steps) as Robot~A executes normally in closed loop, reporting joint $\SR$ as a function of delay $\delta$.

\section{Real-Robot Experiments}
\label{sec:supp:real}

\para{Setup, assets, and runtime.}
\cref{fig:supp:real_assets} presents the physical embodiments, tabletop workspace, and task objects of our heterogeneous real-robot cell.
The platform pairs a 7-DoF Franka FR3 ($8$ active channels; two external + one wrist RGB) facing a dual-arm AgileX Cobot Magic ($2{\times}6$-DoF PiPER arms, $14$ active channels; head + two wrist RGB) operating at $30$\,Hz control and $7.5$\,Hz RGB ingestion ($C{=}2$ chunk $=32$ control steps).
Models are trained via Stage~1$\to$2 (\cref{sec:method:train}) on $80$ synchronized teleoperated demonstration pairs per task ($72$ train / $8$ validation) under the shared $30$-D masked action layout.
Using $4$ video and $8$ action denoising steps with asynchronous parallel decoding, fused QKV projections, compiled RoPE, cached text/history KVs, and CUDA Graphs, a single GPU computes a joint chunk in $0.72$\,s median (vs.\ $3.92$\,s at $25/50$ steps), supporting pause-free closed-loop hardware control.

\begin{table}[b]
\centering\footnotesize\setlength{\tabcolsep}{2.2pt}
\caption{\textbf{Full real-robot joint success rates} ($\SR$ in \%, $95\%$ Wilson interval, and successful trial counts over $20$ trials per task / $60$ total per method) on the heterogeneous Franka\,+\,Cobot team.}
\label{tab:supp:real}
\vspace{-0.45em}
\resizebox{\linewidth}{!}{%
\begin{tabular}{@{}l c c c c@{}}
\toprule
Method & Fruit basket & Toy handoff & Funnel pour & Mean (60 trials) \\
\midrule
\lingbot (per-robot) & 15.0\,[5.2,36.0]\,{\scriptsize(3/20)} & 0.0\,[0.0,16.1]\,{\scriptsize(0/20)} & 25.0\,[11.2,46.9]\,{\scriptsize(5/20)} & 13.3\,[6.9,24.2]\,{\scriptsize(8/60)} \\
\lingbot (joint)     & 0.0\,[0.0,16.1]\,{\scriptsize(0/20)}  & 0.0\,[0.0,16.1]\,{\scriptsize(0/20)} & 0.0\,[0.0,16.1]\,{\scriptsize(0/20)}  & 0.0\,[0.0,6.0]\,{\scriptsize(0/60)} \\
\rowcolor{rowgray}\textbf{\method (ours)} & \textbf{75.0}\,[53.1,88.8]\,{\scriptsize(15/20)} & \textbf{50.0}\,[29.9,70.1]\,{\scriptsize(10/20)} & \textbf{65.0}\,[43.3,81.9]\,{\scriptsize(13/20)} & \textbf{63.3}\,[50.7,74.4]\,{\scriptsize(38/60)} \\
\bottomrule
\end{tabular}}
\end{table}

\begin{figure*}[t]
  \centering
  \includegraphics[width=0.95\textwidth]{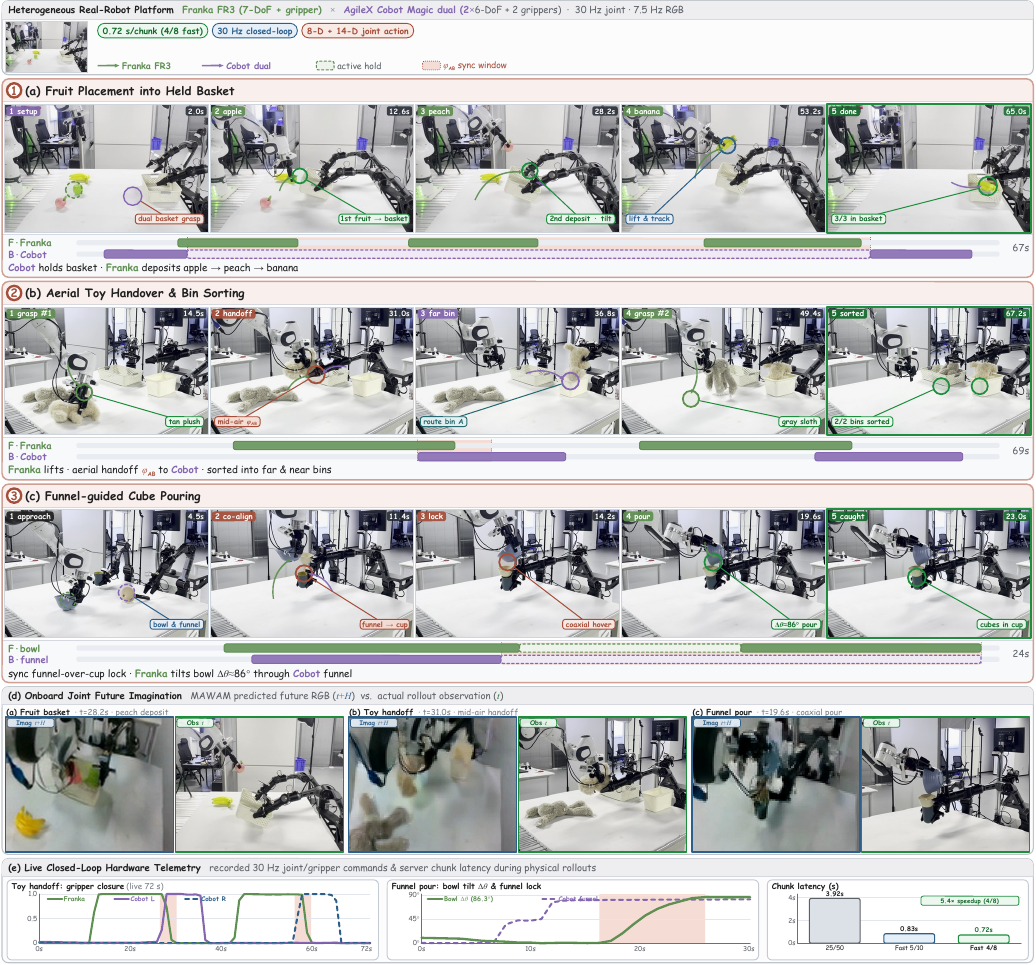}
  \vspace{-0.45em}
  \caption{\textbf{Real-robot collaborative rollouts, onboard joint imagination, and closed-loop telemetry} on the heterogeneous Franka FR3\,+\,AgileX Cobot Magic team.
  \textbf{(a--c)}~Five-frame rollouts and dual-robot Gantt timelines showing active holds and $\varphi_{AB}$ synchronization windows across all three tasks, with measured success rates over $20$ trials per task.
  \textbf{(d)}~Onboard predicted future RGB ($t{+}H$) vs.\ actual rollout observation ($t$) from the unified external side view showing both robots in closed-loop collaboration.
  \textbf{(e)}~Live 30\,Hz closed-loop gripper phase-handoff curves, coaxial funnel lock with $\Delta\theta{\approx}86.3^\circ$ bowl tilt, and accelerated chunk inference latency ($3.92$\,s $\to$ $0.72$\,s, $5.4\times$ speedup).}
  \label{fig:supp:real}
\end{figure*}

\para{Tasks, results, and telemetry.}
The three tasks span distinct physical collaboration regimes (\cref{fig:supp:real}a--c):
(a)~\emph{Fruit basket} (partial-sync, spatial): Cobot holds and tilts a perforated basket while Franka sequentially deposits a green apple, a peach, and a banana bunch;
(b)~\emph{Toy handoff \& sort} (sync $\varphi_{AB}$, spatial): Franka lifts grey and cream plush toys for mid-air handoff to Cobot, which routes each category into its target bin;
(c)~\emph{Funnel cube pour} (sync $SE(3)$, capability): Cobot coaxially stabilizes a handled funnel over a cup while Franka pours wooden cubes from a bowl ($\Delta\theta{\approx}86.3^\circ$).
Across $20$ unseen layouts per task (\cref{tab:supp:real}), \method attains $63.3\%$ mean $\SR$ ($38/60$, $+50.0$~points over per-robot \lingbot, which scores $0/20$ on aerial handoff; flat \lingbot~(joint) scores $0/60$ from $8$-D/$14$-D channel entanglement), consistent with accurate onboard $t{+}H$ imagination and $30$\,Hz phase-locked telemetry (\cref{fig:supp:real}d--e).

\FloatBarrier
\renewcommand{\dbltopfraction}{0.92}
\renewcommand{\dblfloatpagefraction}{0.60}
\section{Additional Qualitative Results on \bench}
\label{sec:supp:more_cases}

\begin{figure*}[p]
  \centering
  \includegraphics[width=\textwidth]{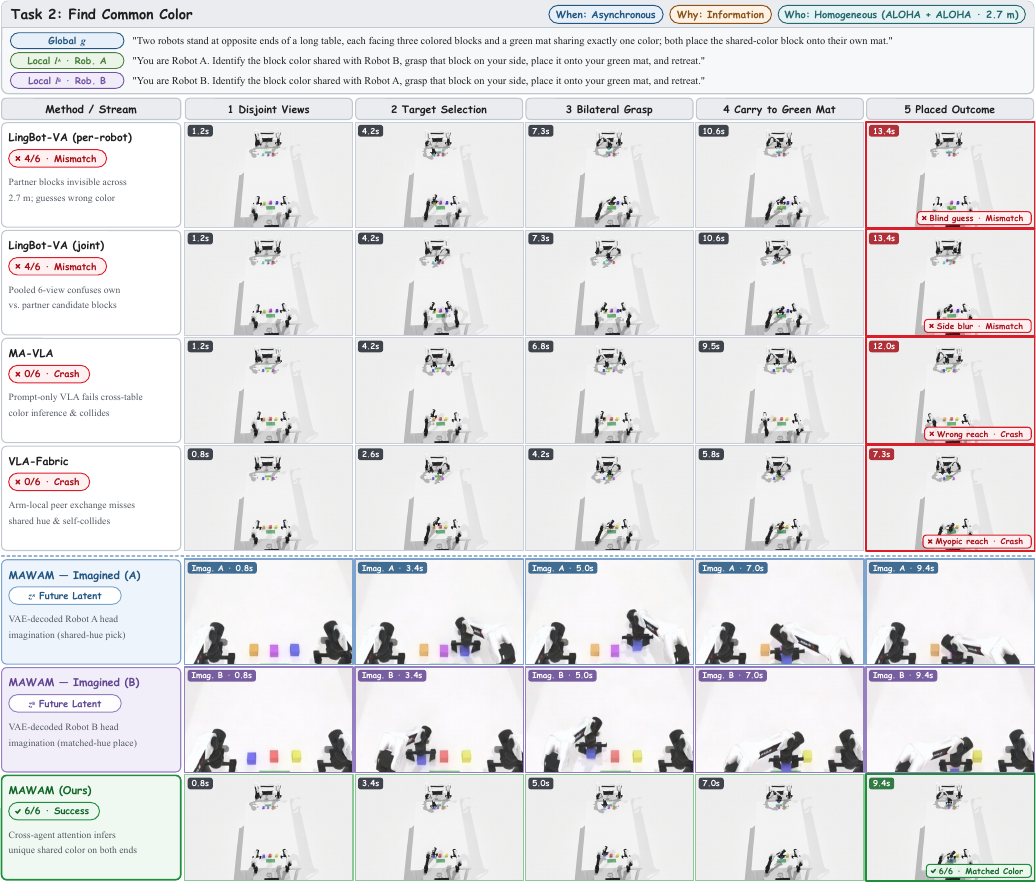}
  \vspace{-0.45em}
  \caption{\textbf{Qualitative comparison on Task~2: \emph{Find common color}} (Asynchronous $\cdot$ Information $\cdot$ Homogeneous ALOHA\,+\,ALOHA across a $2.7$\,m table).
  Decentralized and pooled/local baselines either guess blindly from disjoint views, confuse own vs.\ partner candidate blocks, or fail cross-table reach, whereas \method's cross-agent attention infers the unique shared hue (purple) in both imagined streams and places both matched blocks onto the green mats ($6/6$).}
  \label{fig:supp:case_color}
\end{figure*}

\begin{figure*}[p]
  \centering
  \includegraphics[width=\textwidth]{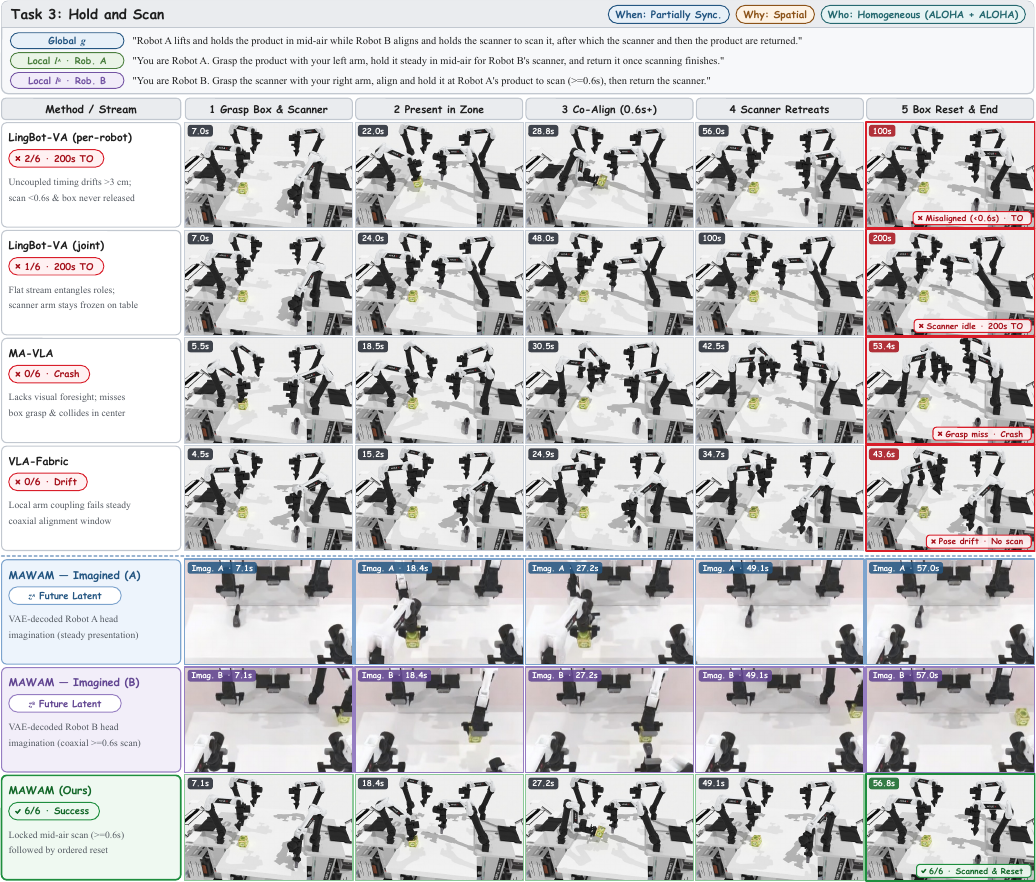}
  \vspace{-0.45em}
  \caption{\textbf{Qualitative comparison on Task~3: \emph{Hold and scan}} (Partially Synchronous $\cdot$ Spatial $\cdot$ Homogeneous ALOHA\,+\,ALOHA).
  Baselines fail to sustain coaxial scanner-to-box alignment for the required ${\ge}0.6$\,s window, freeze the scanner arm, or collide in mid-air.
  \method locks both robots onto a steady mid-air presentation and coaxial scan in $(\hat z^{\,A},\hat z^{\,B})$ before executing an ordered retreat and reset ($6/6$).}
  \label{fig:supp:case_scan}
\end{figure*}

\begin{figure*}[p]
  \centering
  \includegraphics[width=\textwidth]{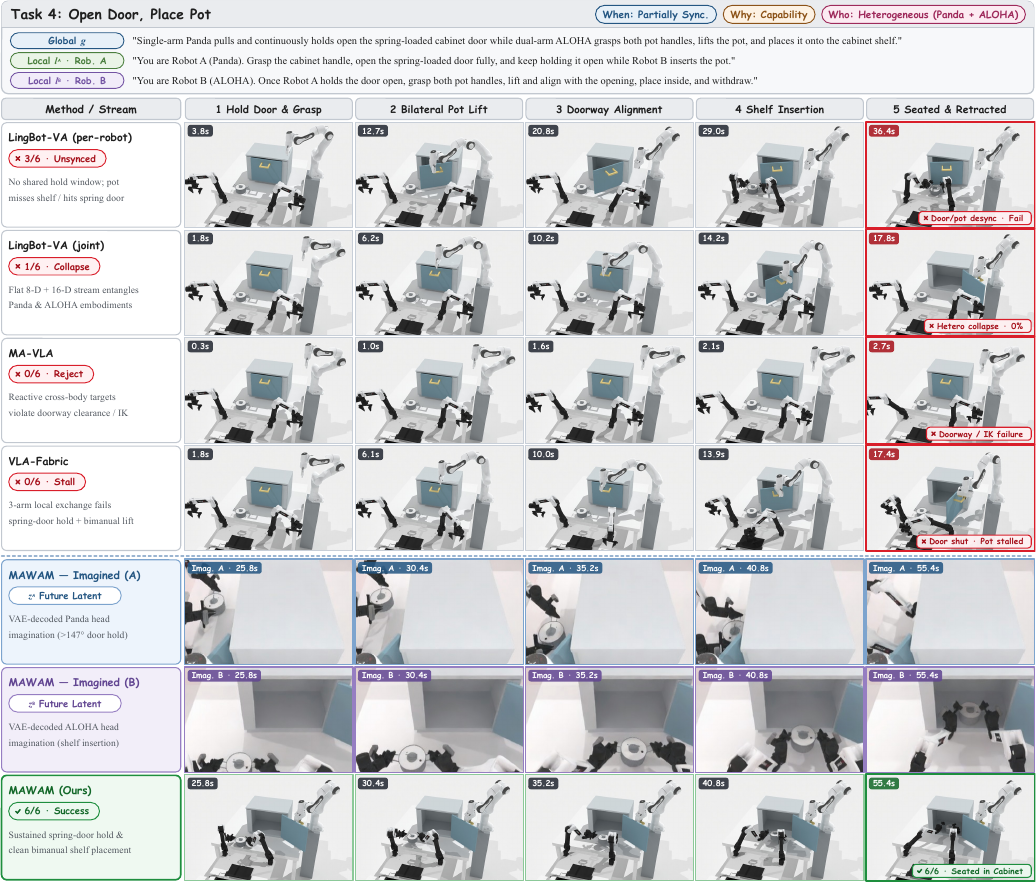}
  \vspace{-0.45em}
  \caption{\textbf{Qualitative comparison on Task~4: \emph{Open door, place pot}} (Partially Synchronous $\cdot$ Capability $\cdot$ Heterogeneous Panda\,+\,ALOHA).
  Decentralized policies desynchronize the spring-loaded door hold from bimanual pot insertion, while flat joint and local-exchange baselines collapse or stall across heterogeneous $8$-D\,+\,$16$-D embodiments.
  \method maintains Panda's $>147^\circ$ door hold while guiding ALOHA's bimanual pot insertion onto the cabinet shelf ($6/6$).}
  \label{fig:supp:case_door}
\end{figure*}

\begin{figure*}[p]
  \centering
  \includegraphics[width=\textwidth]{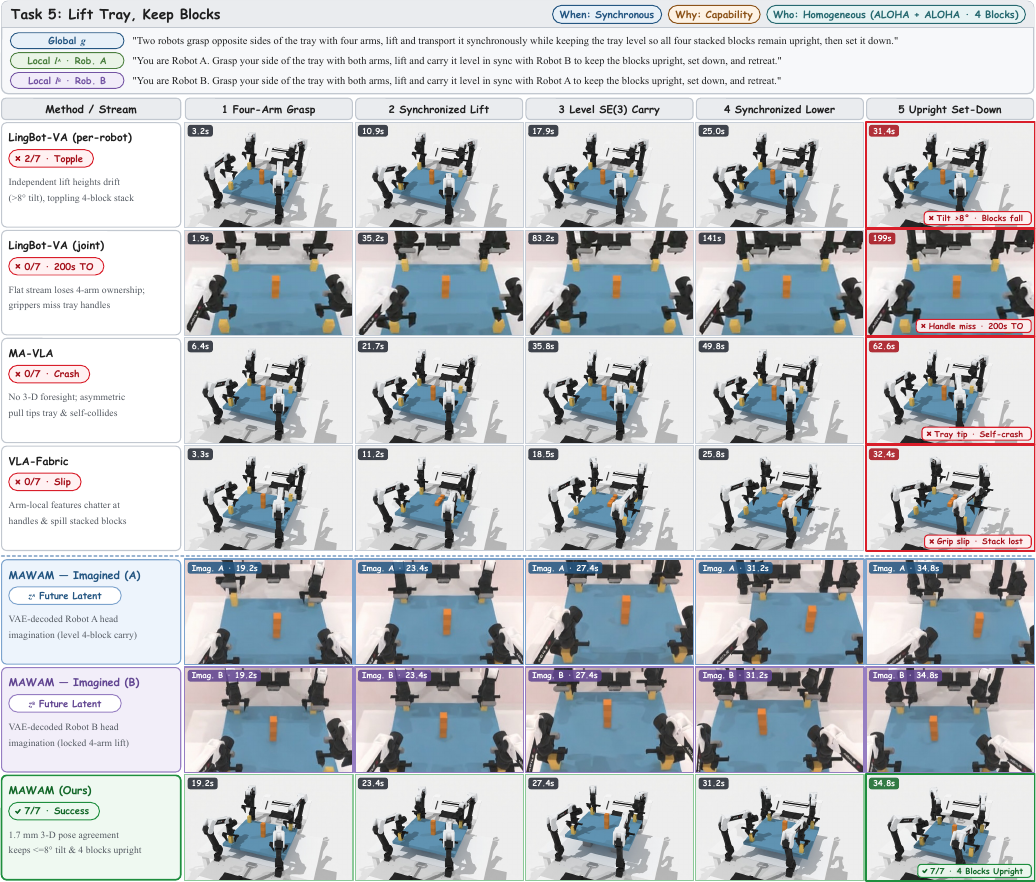}
  \vspace{-0.45em}
  \caption{\textbf{Qualitative comparison on Task~5: \emph{Lift tray, keep blocks}} (Synchronous $\cdot$ Capability $\cdot$ Homogeneous ALOHA\,+\,ALOHA with four stacked blocks).
  Uncoupled or myopic baselines induce closed-chain tilt ($>8^\circ$), miss handles, or slip during four-arm lifting, toppling the four-block stack.
  \method's dual imaginations $(\hat z^{\,A},\hat z^{\,B})$ lock onto a level four-block $SE(3)$ trajectory, keeping all four blocks upright through set-down and retreat ($7/7$).}
  \label{fig:supp:case_tray}
\end{figure*}

\begin{figure*}[p]
  \centering
  \includegraphics[width=\textwidth]{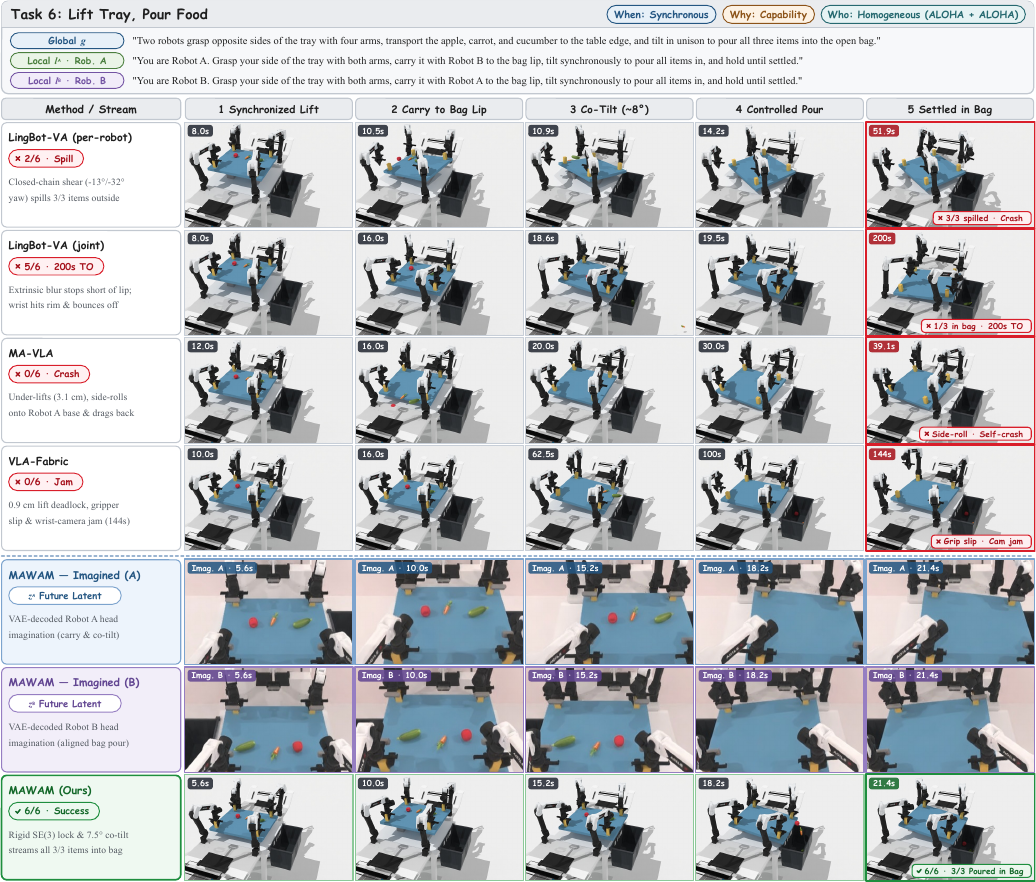}
  \vspace{-0.45em}
  \caption{\textbf{Qualitative comparison on Task~6: \emph{Lift tray, pour food}} (Synchronous $\cdot$ Capability $\cdot$ Homogeneous ALOHA\,+\,ALOHA).
  Baselines shear the closed-chain tray and spill items outside the bag, stall short of the bag lip, or jam grippers during lifting.
  \method coordinates a level transport to the bag lip followed by a synchronized ${\sim}7.5^\circ$ co-tilt that pours all $3/3$ food items cleanly into the bag ($6/6$).}
  \label{fig:supp:case_food}
\end{figure*}

\begin{figure*}[p]
  \centering
  \includegraphics[width=\textwidth]{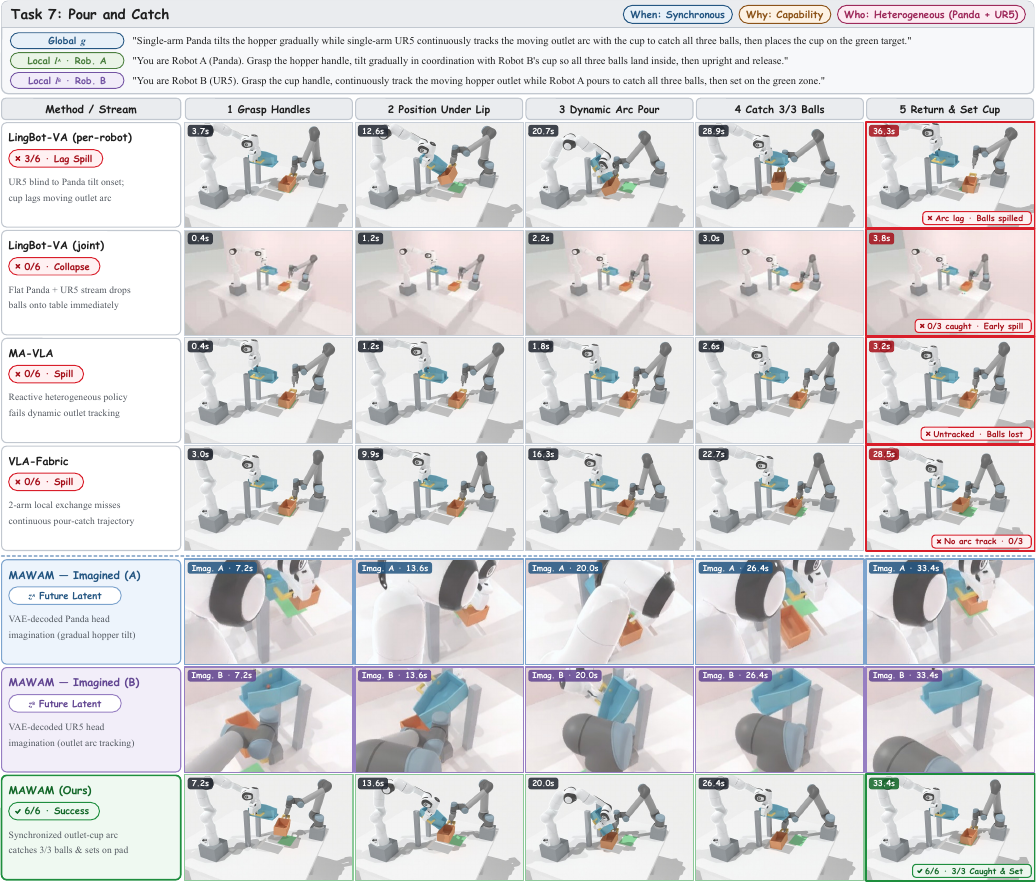}
  \vspace{-0.45em}
  \caption{\textbf{Qualitative comparison on Task~7: \emph{Pour and catch}} (Synchronous $\cdot$ Capability $\cdot$ Heterogeneous Panda\,+\,UR5).
  Without shared visual foresight across heterogeneous arms, baselines either lag behind Panda's moving hopper outlet arc or spill the balls prematurely onto the table.
  \method couples Panda's gradual hopper tilt with UR5's continuous outlet-arc tracking to catch all $3/3$ balls and set the cup onto the target pad ($6/6$).}
  \label{fig:supp:case_balls}
\end{figure*}

To complement the selected qualitative strip in \cref{fig:qual_dual}, \cref{fig:supp:case_bowl,fig:supp:case_color,fig:supp:case_scan,fig:supp:case_door,fig:supp:case_tray,fig:supp:case_food,fig:supp:case_balls} present complete five-stage rollout comparisons across all seven \bench tasks.
At the top of each showcase table, we list the task's taxonomy coordinates (\emph{when}, \emph{why}, \emph{who}) together with the exact global team prompt $g$ and per-robot local role prompts $(\ell^A,\ell^B)$ supplied at evaluation time.
Rows~1--4 display representative failure trajectories of the four multi-robot paradigms from \cref{fig:qual_dual}—decentralized \lingbot~(per-robot), flat-concatenation \lingbot~(joint), centralized action-only MA-VLA, and arm-level peer-exchange VLA-Fabric—highlighting each paradigm's characteristic coordination breakdown (blind early entry, cross-embodiment or cross-arm role entanglement, reactive collision without visual foresight, and myopic arm-local drift).
Directly above \method's execution row, Rows~5--6 visualize \method's synchronized future head-camera latents $(\hat z^{\,A},\hat z^{\,B})$ decoded offline via the frozen causal Wan2.2-VAE at the exact timestamps of \method's closed-loop execution keyframes (Row~7), showing how shared future visual foresight aligns both robots across asynchronous handoff, cross-table information inference, partially synchronous holding, and tightly coupled $SE(3)$ / dynamic co-manipulation.

\begin{strip}
  \vspace{-0.8em}
  \centering
  \includegraphics[width=0.94\textwidth]{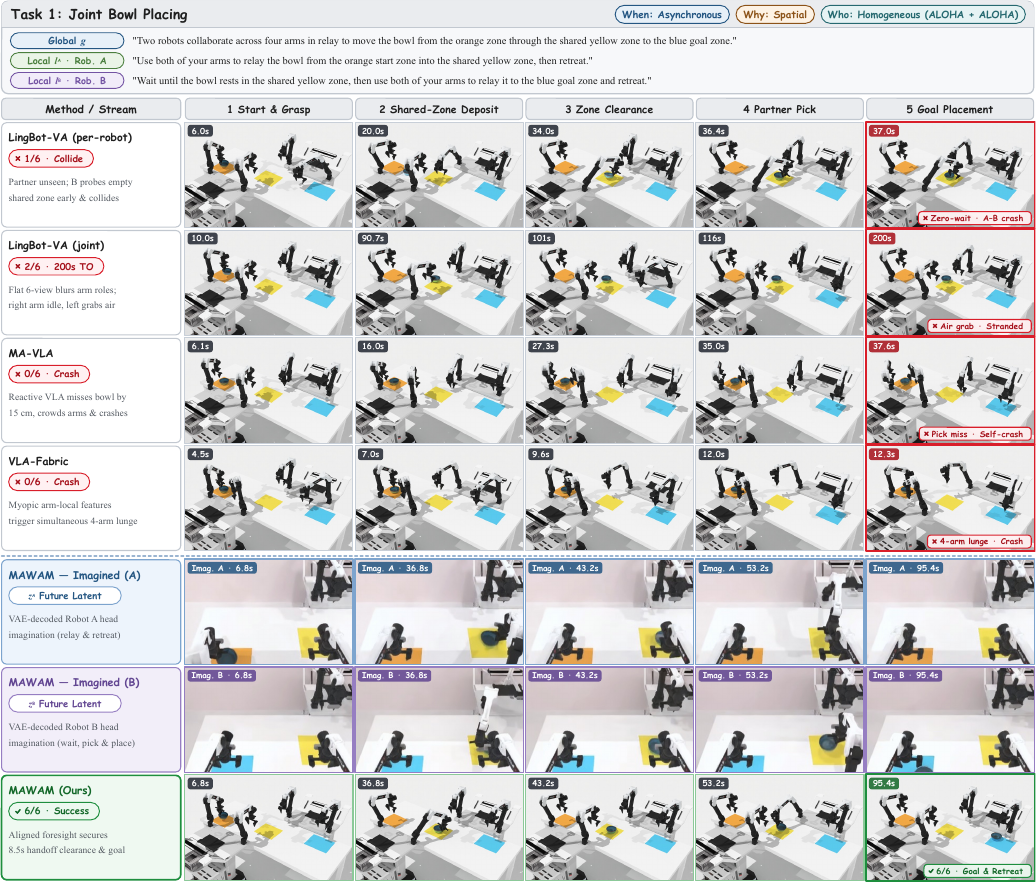}
  \vspace{-0.5em}
  \captionsetup{hypcap=false}
  \captionof{figure}{\textbf{Qualitative comparison on Task~1: \emph{Joint bowl placing}} (Asynchronous $\cdot$ Spatial $\cdot$ Homogeneous ALOHA\,+\,ALOHA).
  Baselines collide in the shared zone without partner awareness (\lingbot per-robot), lose arm ownership under flat view pooling (\lingbot joint), or lunge reactively into multi-arm collisions (MA-VLA, VLA-Fabric).
  \method's synchronized imaginations $(\hat z^{\,A},\hat z^{\,B})$ anticipate zone clearance before partner entry, completing all $6/6$ milestones.}
  \captionsetup{hypcap=true}
  \label{fig:supp:case_bowl}
  \vspace{-1.2em}
\end{strip}

\end{document}